\documentclass[fleqn,10pt]{wlscirep}
\usepackage[utf8]{inputenc}
\usepackage[T1]{fontenc}
\usepackage{multirow,bigstrut}
\usepackage{makecell, caption, booktabs}
\usepackage{array}

\usepackage{adjustbox}
\usepackage{graphicx} 
\usepackage{gensymb}
\usepackage{siunitx}
\usepackage{todonotes}
\usepackage{algorithm}
\usepackage{algpseudocode}
\usepackage{amsmath}
\usepackage{amsfonts}
\usepackage{booktabs}
\usepackage[table]{xcolor}
\definecolor{lightgray}{gray}{0.9}
\usepackage{subcaption, geometry}
\usepackage{stackengine}
\usepackage{amssymb}
\usepackage{pifont}
\newcommand{\cmark}{\ding{51}}%
\newcommand{\xmark}{\ding{55}}%

\newcommand{\omitme}[1]{}
\newcommand{\addref}[1]{} 
\newcommand{\ourloss}[1]{Cur.}
\newcommand{\td}[1]{\textcolor{red}{#1}}

\title{CHMv2: Improvements in Global Canopy Height Mapping using DINOv3}

\author[1,+*]{John Brandt}
\author[2,+]{Seungeun Yi}
\author[3,+]{Jamie Tolan}
\author[4]{Xinyuan Li}
\author[1]{Peter Potapov}
\author[1]{Jessica~Ertel}
\author[1]{Justine Spore}
\author[2]{Huy V. Vo}
\author[2]{Micha\"{e}l Ramamonjisoa}
\author[2]{Patrick Labatut}
\author[2]{Piotr~Bojanowski}
\author[2+]{Camille Couprie}
\affil[1]{World Resources Institute, 10 G St NE \#800, Washington, DC 20002, USA}
\affil[2]{Fundamental AI Research (FAIR), Meta, 75002 Paris, France}
\affil[3]{Meta, 1 Hacker Way, Menlo Park, CA 94025, USA}
\affil[4]{University of Maryland, Department of Geography, College Park, MD 20742, USA}

\affil[*]{John.Brandt@wri.org}
\affil[+]{these authors contributed equally to this work}

\begin{abstract}
Accurate canopy height information is essential for quantifying forest carbon, monitoring restoration and degradation, and assessing habitat structure, yet high-fidelity measurements from airborne laser scanning (ALS) remain unevenly available globally.  Here we present CHMv2, a global, meter-resolution canopy height map derived from high-resolution optical satellite imagery using a depth-estimation model built on DINOv3 and trained against ALS canopy height models. Compared to existing products, CHMv2 substantially improves accuracy, reduces bias in tall forests, and better preserves fine-scale structure such as canopy edges and gaps. These gains are enabled by a large expansion of geographically diverse training data, automated data curation and registration, and a loss formulation and data sampling strategy tailored to canopy height distributions. We validate CHMv2 against independent ALS test sets and against tens of millions of GEDI and ICESat-2 observations, demonstrating consistent performance across major forest biomes.
\end{abstract}

\begin{document}

\flushbottom

\maketitle


    


\section{Background \& Summary}

\begin{figure}[ht]
    \begin{center}
    \includegraphics[width=.95\linewidth]
    {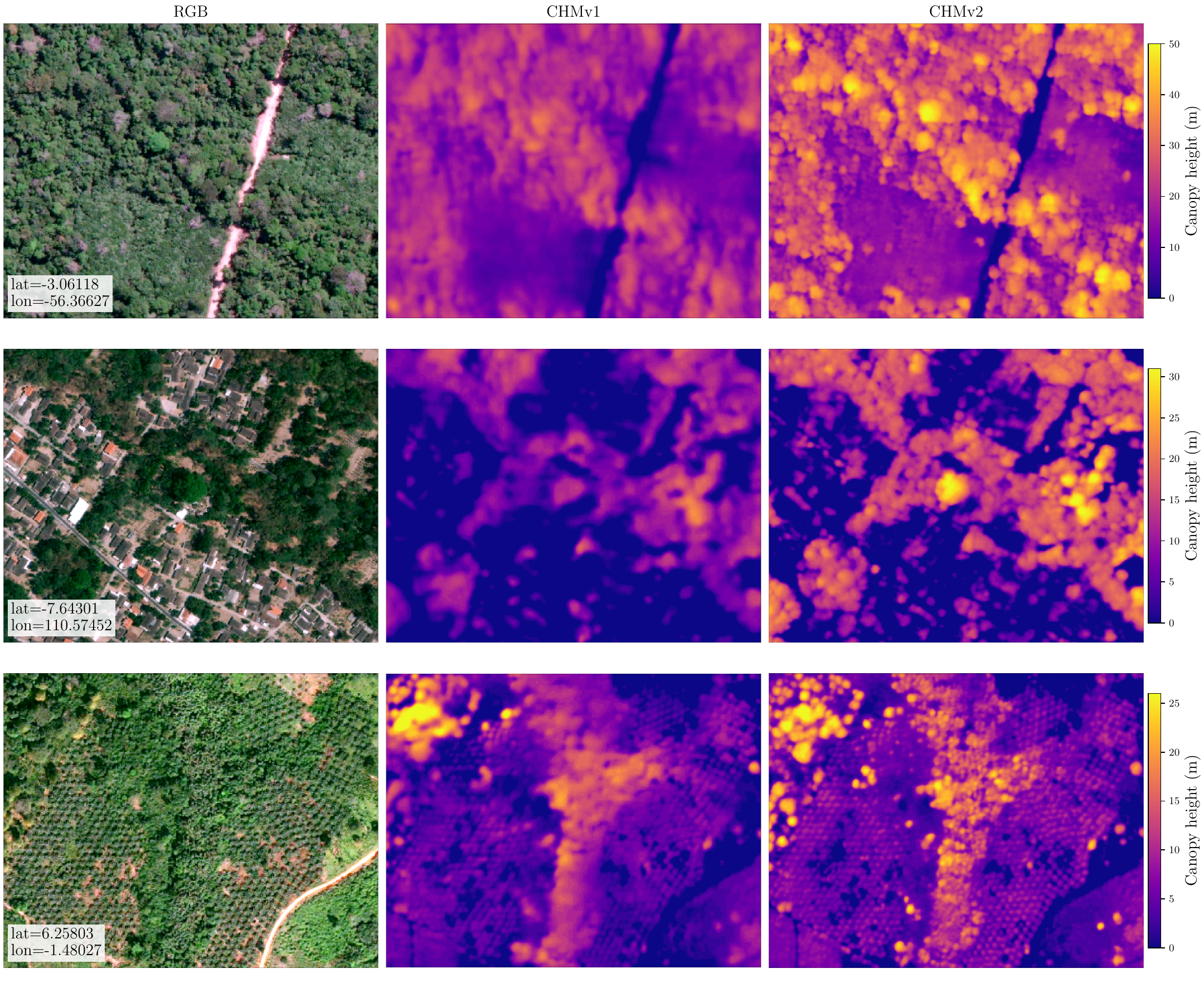}
    \end{center}
    \caption{Visual improvements from CHMv1 to CHMv2 in a disturbance area in the Amazon (top), an urban forest in Central Java, Indonesia (middle) and a plantation in the Ghanaian cocoa belt (bottom). 
}
    \label{fig:teaser}
\end{figure}


Remote sensing assessments of canopy height and structure are important for forest land management, including the monitoring of forest and landscape restoration\cite{ALMEIDA2019192}, detecting forest degradation and regrowth\cite{Senf2020}, and measuring above-ground biomass\cite{ZHANG201444}. Wall-to-wall maps of canopy height provide a bridge between field-based measurements and landscape or national scale land management decisions and are increasingly needed to support climate mitigation accounting, restoration planning, biodiversity assessments, and emerging digital monitoring, reporting, and verification (dMRV) approaches for climate finance\cite{Brandt2025}.
Remote-sensing derived canopy height datasets have been used to improve land cover mapping\cite{YUH2023101955}, measure the carbon benefits of afforestation\cite{doi:10.1073/pnas.2304988120}, model forest fire fuel\cite{GALE2021112282}, and estimate aboveground biomass\cite{TAMIMINIA2024102404} among other applications. Beyond percentile-based canopy height estimates (e.g., p95), fine-scale structural metrics such as gap fraction, edge density, and height heterogeneity are typically derived from ALS-based canopy height models and are not recoverable from medium-resolution products. In forest monitoring contexts, these structural metrics are commonly used as covariates or stratification layers to reduce uncertainty in biomass estimation, support sampling design for field verification, and characterize heterogeneous systems such as agroforestry and secondary forests, where canopy structure varies at sub-hectare scales\cite{Haneda2026, rs8010021}. Extending access to such structural information beyond regions with ALS coverage remains a key challenge for global monitoring.

Global canopy height datasets derived from observation satellite data have utilized a variety of methodological approaches, typically involving a fusion of optical satellite imagery with LiDAR observations derived from spaceborne instruments like Global Ecosystem Dynamics Investigation (GEDI)\cite{gedil1b} and ICESat-2\cite{IceSatHeight} or airborne laser scanning\cite{Balestra2024}. Among globally available datasets, Potapov \textit{et al.}\cite{Potapov2021Mapping} fused GEDI RH95 (percentile of relative height) data with Landsat multispectral data using a locally calibrated regression tree ensemble algorithm to generate a 30-meter global canopy height product for 2020. Using Sentinel-2 data, Lang \textit{et al.}\cite{Lang2022High} applied an ensemble of convolutional neural network (CNN) models to  predict the GEDI RH98 value for each 10-meter pixel. Despite the diversity in methods for canopy height mapping, these datasets share three limitations. First, they have difficulty capturing short or low-stature vegetation\cite{Hunter2025}. Second, they are not high enough resolution to map forest structure or complexity which is a key input to degradation, restoration, and biodiversity monitoring\cite{https://doi.org/10.1111/ddi.13644}. Third, the majority of global-scale, medium resolution height products rely solely on spaceborne LiDAR from GEDI or ICESat-2, rather than airborne LiDAR, which may limit their applicability for land management at granular spatial scales where the gold standard evaluation criteria is agreement with airborne LiDAR data\cite{gca}.

Recent breakthroughs in vision transformers (ViTs) and self-supervised learning (SSL) have substantially improved performance in dense prediction tasks, including monocular depth estimation. Self-supervised methods such as DINOv2 \cite{Oquab2023SSL} and more recently DINOv3 \cite{simeoni2025dinov3} have demonstrated strong capabilities in extracting semantically meaningful representations from large corpora of unlabeled imagery. These representations have shown improved generalizability across different domains, enabling DINOv3 to establish state of the art results on a variety of image tasks within earth observation. For the task of canopy height mapping, the DINOv3 SSL features show clear accuracy improvements over strong baselines when measured on the Open-Canopy benchmark \cite{fogel2025open}. Features from SSL ViTs have also demonstrated improved generalization to new geographies, compared to U-Net models\cite{10.1007/978-3-319-24574-4_28}, for Landsat-based canopy height mapping \cite{10640630}.

While ALS provides high fidelity 3D structural information, its global availability is highly uneven. One recent review\cite{Balestra2024} found that 90\% of studies fusing LiDAR data with earth observation data were focused in North America, Europe, and Asia. Because ALS coverage is uneven, models trained from ALS alone can inherit strong geographic priors and under-perform in underrepresented forest types. SSL representations provide a complementary signal by learning broad visual semantics and structure-related features from diverse imagery, improving robustness when supervised ALS training data are sparse or regionally biased. Moreover, ALS acquisitions are typically episodic in time, whereas archived RGB imagery enables retrospective canopy height estimation for historical baselines and change analyses in areas without repeated ALS coverage. While initiatives such as the Global Canopy Atlas\cite{gca} may alleviate the data coverage limitation, there remain significant operational challenges to fusing ALS data with optical imagery. Pairing ALS-derived canopy height models with optical imagery introduces several sources of noise. ALS and optical acquisitions often differ in date by months to years, leading to mismatched phenology or land use change between inputs and targets. Even when acquisitions are temporally close, geolocation error and viewing-geometry differences can induce local misregistration. Because ALS-derived CHMs provide fine-scale structural supervision, this misregistration disproportionately degrades the learning of detailed canopy structure. These challenges motivate automated data cleaning and registration procedures that can scale to millions of training pairs.

Tolan \textit{et al.}\cite{tolan2024very} developed the first high resolution map of global canopy height (hereafter referred to as CHMv1) by training a depth estimation model using a frozen DINOv2\cite{Oquab2023SSL} backbone as a fixed feature extractor, which reduces the amount of task-specific training required, with high resolution optical input and ALS-derived CHM output. The ALS data consisted of sparse LiDAR transects within the USA from NEON \cite{neonals} and the model was applied globally with a low-resolution correction factor applied via a secondary CNN trained on GEDI footprints.  The quality of this map varies widely across spatial scales, reference canopy height, and degree of heterogeneity in canopy structure. In a large-scale comparison to 3,458 ALS data reference transects, Fischer \textit{et al.}\cite{gca} found that while CHMv1\cite{tolan2024very} compares favorably to medium-resolution global products, residual errors can still be limiting for sub-landscape analyses of canopy height variability. These results are consistent with the conclusions of Moudr\'y \textit{et al.}\cite{moudry}, which found that CHMv1\cite{tolan2024very} had significant underestimation of canopy height across three biodiversity areas in California, New Zealand, and Switzerland. Within a sparser forest system, D'Amico \textit{et al.}\cite{DAMICO2025127197} found that CHMv1 had comparably high accuracy for mapping canopy height of agroforestry tree systems in Italy. 

Together, these limitations highlight the need for a globally consistent, meter-resolution canopy height dataset trained on more representative ALS coverage and supported by modern SSL backbones. Existing global products either lack high spatial resolution, rely on LiDAR supervision concentrated in a few regions, or exhibit substantial geographic bias. To address this gap, we develop a new global 1-meter canopy height dataset that builds directly on the CHMv1\cite{tolan2024very} framework but incorporates several key advances. We replace the DINOv2-H encoder with the more capable DINOv3 Sat-L backbone, expand and rigorously clean a geographically diverse ALS training corpus, and apply improved RGB-CHM registration to reduce label noise. We further introduce a loss formulation tailored to canopy height distributions and structural variability. Through ablations and evaluations across multiple benchmarks, we show that these improvements substantially reduce bias, increase accuracy across canopy height ranges, and enhance fine-scale structural fidelity relative to previous global products (Figure \ref{fig:teaser}). 

\section{Methods}
\subsection{Overview of Workflow} 
This paper presents a global 1-meter canopy height map generated from high-resolution satellite imagery with a depth estimation model trained against ALS data. The imagery source is the same as in CHMv1\cite{tolan2024very}, with approximately 80\% of images spanning 2018 to 2020. Compared to CHMv1 \cite{tolan2024very}, we substantially expand and clean the training dataset, improve the model backbone and decoder, and identify an improved training curriculum (loss, dataset mix, optimization parameters). These differences yield substantial qualitative and quantitative accuracy improvements, which are confirmed via rigorous ablations and validation analyses. 

\subsection{Input Imagery} 
The satellite imagery resolution, acquisition dates, and preprocessing are discussed in detail in the CHMv1 manuscript\cite{tolan2024very}. In summary, we utilize Maxar Vivid2 mosaic imagery as input imagery for model training and inference. This dataset mosaics together imagery from multiple instruments (WorldView-2, WorldView-3, and Quickbird II) and observation dates, with a pixel size of 0.597 m GSD at the equator.

\subsection{Decoder training and validation datasets}

\begin{figure}[htb]
    \begin{center}
    \includegraphics[width=.95\linewidth]
    {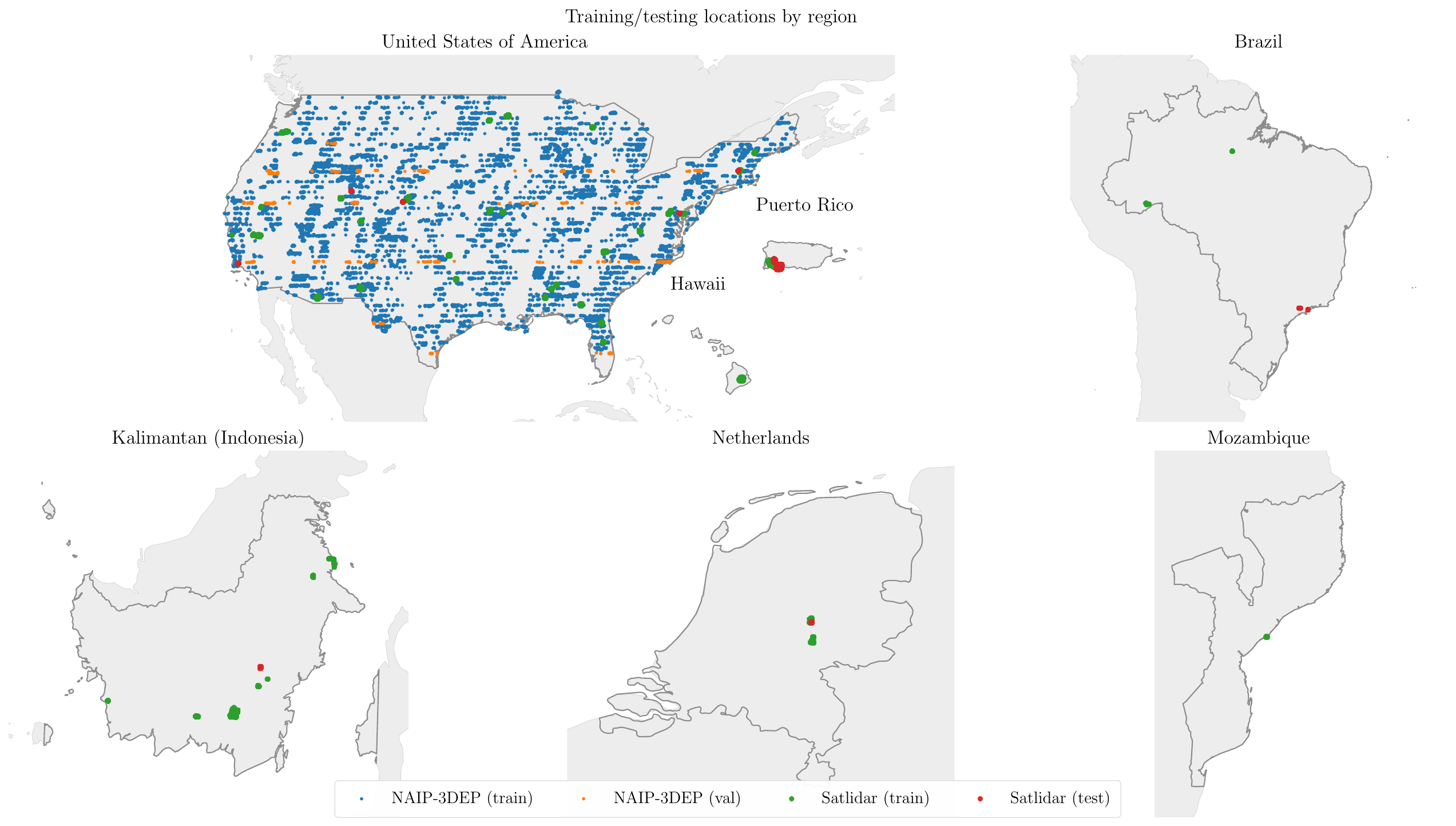}
    \end{center}
    \caption{
        Training and testing/validation data locations for NAIP-3DEP and SatLidar datasets.
    }
    \label{fig:dataloc}
\end{figure}

\subsubsection{NAIP-3DEP dataset}\label{sec:naip}

The 3D Elevation Program (3DEP) \cite{opentopo3dep} collects extensive ALS and topographic data across the United States. Allred \textit{et al.}\cite{allred2025canopy} curated a dataset of approximately 22 million paired 3DEP ALS and NAIP optical chips across the contiguous United States with a 256 $\times$ 256 meter chip size. We created a derivative dataset (NAIP-3DEP) based on post-processing, cleaning, and sampling of the Allred \textit{et al.}\cite{allred2025canopy} dataset. First, we selected approximately 360k NAIP-CHM pairs with CHM and NAIP acquisition dates within 60 days of each other. Next, building footprints were set to zero canopy height using the Microsoft US Building Footprint dataset\cite{microsoft2018usbuildings}. Approximately 60k CHM tiles with striping, leaf-off acquisition, or evident land use change between input and output were then discarded with an automated method described in Section~\ref{sec:dataclean}. Finally, we performed automatic registration of the optical and ALS data as outlined in Section~\ref{sec:datareg}. The remaining samples, about 300k,  were split with geographic stratification into approximately 280k training and 20k validation patches. The dataset consists of 427 $\times$ 427 NAIP RGB optical images at 0.6 m GSD and ALS-derived CHM at the same pixel resolution but with a 1 m GSD. The CHM data is sampled to 0.6 m GSD with a super-resolution model based on DSen2\cite{Lanaras_2018}. The locations of the train and validation data for NAIP-3DEP are displayed in Figure ~\ref{fig:dataloc}.

\subsubsection{SatLidar dataset}\label{sec:satlidar}
We construct SatLidar v2, a curated and registered revision of the SatLidar v1 dataset introduced in the DINOv3\cite{simeoni2025dinov3} paper. SatLidar v1 consists of approximately one million $512\times 512$ images with ALS ground truths split into train/val/test splits with ratios 8/1/1. The splits include the Neon dataset used by CHMv1\cite{tolan2024very}. SatLidar v2 corresponds to a curated and registered version of SatLidar v1 where poor quality training samples have been discarded (Section~\ref{sec:dataclean}) and input and output datasets have been registered (Section~\ref{sec:datareg}). It includes approximately 726K train, 89K val and 91K test samples. The locations of the train, val, and test data for SatLidar v2 are displayed in Figure~\ref{fig:dataloc}.

\subsubsection{Multisource dataset}\label{sec:multisource}
The multisource dataset is a test dataset consisting of 433 triplets of images from three different sources (Maxar, Maxar taken at a different date, and NAIP) with associated ALS ground truths. All images are located in the `WLOU' NEON site (Colorado), with ALS observations acquired in 2019~\cite{NeonData}. Similar to the other datasets, the multisource dataset was curated following classification with a linear probe, and the same alignment procedure we used for SatLidar v2. 

\subsubsection{NAIP Sea dataset}
The datasets described in Sections~\ref{sec:naip} to ~\ref{sec:multisource} have spatial footprints determined by the locations of ALS transects, which are sparse or non-existent over open water. However, choppy water can look quite similar to forests with snow on the ground in high resolution imagery leading to erroneous height predictions over water. To remedy this, we randomly sample open water and shorelines within the United States in NAIP imagery based on the Global Lakes and Wetlands Database\cite{LEHNER20041} and create canopy height target values of zero for approximately 3,500 samples. This dataset is included in our data mixes to improve data fidelity over water.

\subsection{Training data curation}

\subsubsection{Data cleaning}\label{sec:dataclean}

We used DINOv3 as a few-shot anomaly detector to discard poor quality pairs in the training datasets \cite{zhai2025foundationvisualencoderssecretly}. A weak CHM model, trained on approximately 5k manually reviewed samples, produced preliminary RGB / CHM label / CHM prediction triptychs. We manually annotated "keep" and "discard" labels covering land-use mismatch and CHM sensor artifacts for a small subset ($\approx$10k) of our data stratified by tree cover and height metrics based on visual comparison of the input image, the weak prediction, and the label. A linear probe was trained on the L1 difference between frozen DINOv3 embeddings (prediction - label) for these samples. The linear probe was applied to all the training corpus and samples with below a 0.5 keep score were removed based on the receiver operating characteristic, excluding about 15\% of all data. All removals in the validation and test set were manually visually inspected to avoid erroneous removal of difficult validation or test samples.

\begin{figure}[ht!]
    \begin{subfigure}[b]{0.95\textwidth}
    \centering
    \includegraphics[width=.8\linewidth]{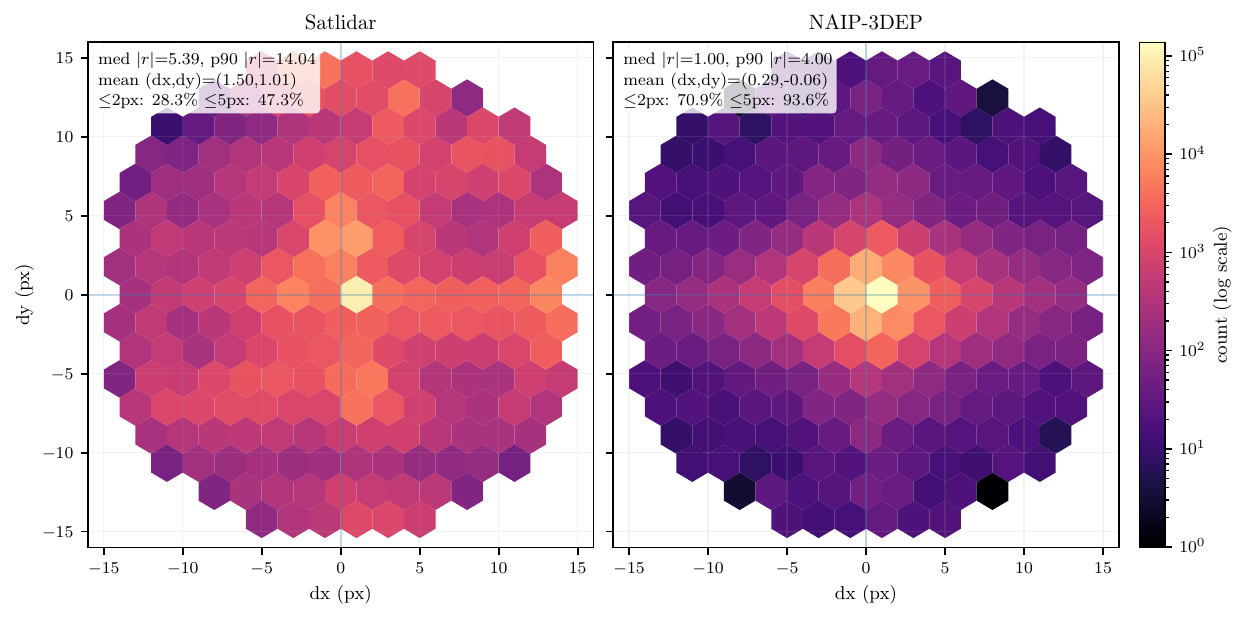}
    \caption{
        Global best shift magnitude for SatLidar and NAIP-3DEP. NAIP imagery is much better registered, as is the ALS data. SatLidar has many input data sources and satellites, and the raw data is much more poorly aligned.
    }
    \label{fig:cleaningfig}
    \end{subfigure}\\
    \begin{subfigure}[b]{0.95\textwidth}
    \centering
    \includegraphics[width=.8\linewidth]
    {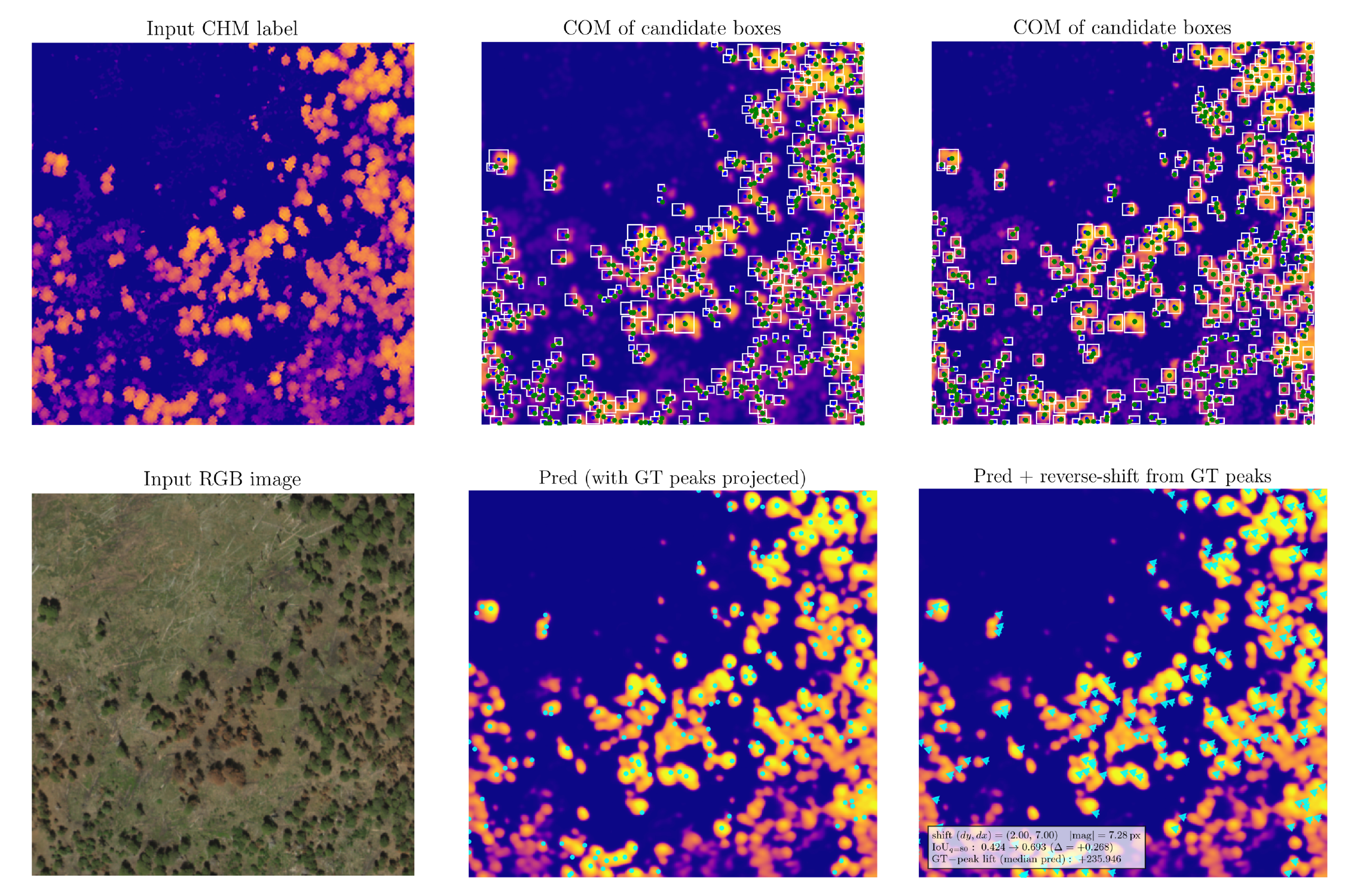}
    \caption{
        Summary of our alignment methodology for local alignment (top) with bounding box proxies and global alignment with weak predictions (bottom). 
    }
    \label{fig:satfig}
    \end{subfigure}
\caption{Data registration methodology and analysis.}
\end{figure}

\subsubsection{Data registration}\label{sec:datareg}
Accurate spatial alignment of RGB imagery and CHM labels is essential for high quality canopy height mapping. Recent monocular depth estimation approaches, such as Depth Anything v2 \cite{yang2024depthv2}, have found that models trained on synthetic data generate sharper depth predictions than models trained on non-synthetic data. Because optical and ALS data are collected at different times by different sensors, such paired datasets have large amounts of random misalignment due to geolocation and differences in viewing geometries. Without correcting for these misalignments, there is no consistent supervisory signal for detailed local 3D structure. Despite the importance of well aligned RGB and CHM data, reliable cross-domain registration remains challenging in forested systems. Many cross-domain registration pipelines for earth observation rely on photometric consistency, repeatable keypoints, or significant mutual information between domains\cite{LI2023103574, rs17040646}. These assumptions are infrequently met for canopy height mapping datasets because reflectance is not a consistent proxy for height (limiting mutual cross domain information) and forest structure, which is dominated by textured crowns, shadows, and view-illumination effects, is not a reliable source for keypoint identification.

A further complication is that the most reliable way to register images of different domains (CHM labels and RGB imagery) is to compare like with like via domain adaptation \cite{MAHAPATRA2020107109, 10320402}. However, the practical application of domain adaptation to transforming optical images to a canopy-height representation is made difficult by the varying amount of mutual information between these domains. We found that supervised approaches were necessary for generalization, but such approaches require a reasonably accurate CHM prediction model for domain adaptation, which itself must be trained using well-aligned RGB--CHM pairs. If the initial supervision is misaligned, the model converges to blurred and conservative predictions that are unsuitable for reliable registration. To break this chicken-and-egg dependency, we bootstrap alignment using auxiliary structural cues from an independent tree detection model together with a weak CHM model trained on a carefully cleaned subset of the data (Figure~\ref{fig:cleaningfig}). 

We correct local misalignment using tree detections as control points. A DINO DETR\cite{zhang2022dinodetrimproveddenoising} detector trained on human annotations from MillionTrees\cite{MillionTrees2025} produces bounding boxes for individual trees for each input RGB image. Because these tree predictions are derived from a model trained on manually annotated images, they do not exhibit misalignment between input and label. Within each predicted tree box, we compute one or more canopy height "centers of mass" (COMs). The per-tree offsets from box center to COM are clustered with DBSCAN\cite{ester1996density}, regularized with per-cluster medians to resolve disagreements, and interpolated to a dense warp field using a thin-plate radial basis function. Alignment is performed iteratively: offsets are estimated, a small warp is composed into a velocity field, and CHM predictions are re-measured until the incremental offset is smaller than 1 pixel.

We correct global misalignment by estimating a rigid 2D translation between predictions and labels with Fast Fourier Transform (FFT)-based cross-correlation\cite{Kidorf1984PracticalFF}. For each tile, we first apply a lightweight "weak" canopy height model, trained on data with local misalignments corrected via the detection-based control points, to both the satellite imagery and the ALS CHM to place both inputs and labels in a common radiometric space suitable for automated alignment. We then detect high-canopy peaks and compute an initial alignment via FFT-based cross-correlation of peak masks, followed by a small local grid search for refinement. Candidate shifts are scored with IoU of high-canopy regions and a height "lift" term calculated as the median delta in canopy height of the shifted prediction at each ground truth peak. The shifts are only applied when they improve IoU or height lift related to a zero-shift baseline. Different steps of the registration process are illustrated in Figure~\ref{fig:satfig}.

The final training dataset is composed of the original optical data with CHM data that has been first aligned globally with a per-tile offset, and then locally with a dense warp field, to the optical data. Overall, 85\% of training samples with non-zero canopy height targets benefit from the alignment. The SatLidar dataset had much more severe misalignment  (53\% of samples with $\geq$ 5 px shift) than did the NAIP-3DEP dataset (6.4\% of samples with $\geq$ 5 px shift) (Figure~\ref{fig:cleaningfig}). The alignment process improved the average IoU of the "weak" canopy height model for the per-tile p80 canopy height from 0.31 to 0.42.



\subsection{Decoder loss functions}
Canopy height mapping differs from conventional monocular depth estimation in both viewing geometry and depth statistics. Typical monocular depth tasks are trained on ground-based or oblique camera views, emphasize near-camera depth accuracy, and contain few zero-depth regions. In contrast, canopy height estimation is performed from a nadir, overhead viewing angle and includes large areas of zero or near-zero height, while simultaneously requiring accurate prediction of tall, spatially sparse structures. These characteristics alter the distribution of prediction errors and the relative importance of different error modes. We therefore evaluate several loss formulations with regards to MAE (Mean Absolute Error), block-$R^2$ with $50\times50$ pixels blocks, bias, and high depth ($\ge 30$ meter) mean bias error.

\paragraph{SiLog loss.} We evaluate the SiLog loss\cite{eigen2014depthmappredictionsingle} function commonly used in depth estimation approaches such as DepthAnythingv2\cite{yang2024depthv2}, DepthPro\cite{bochkovskii2025depthprosharpmonocular}, and CHMv1\cite{tolan2024very}. We find that training with SiLog loss alone for the task of canopy height mapping is insufficient because canopy height maps have a significant proportion of zero or near-zero values, which can dominate the objective in log space and reduce emphasis on errors in tall, spatially sparse canopy, leading to negative bias at high canopy height.

\paragraph{Charbonnier loss.} Because training with SiLog loss resulted in large high-depth mean bias error, we evaluate model results when training with Charbonnier loss\cite{charbonnier}, a smooth variant of the L1 loss. These models converge to trivial solutions by predicting the batch mean value. As such, we consider the effect of starting model training with SiLog loss to establish relative depth structure and gradually transitioning to Charbonnier loss during training to encourage the model to have correct predictions in linear space. 

\paragraph{Gradient loss.} When training depth estimation models with L1 variants, it is common to utilize a gradient loss with Sobel or Laplacian operators to penalize deviations in spatial structure \cite{yang2024depthv2, bochkovskii2025depthprosharpmonocular, Jiao_2018_ECCV}. Notably, MiDaS \cite{midas} introduced a multi-scale, scale- and shift-invariant gradient loss that operates in log-depth space and is evaluated across multiple spatial resolutions of both the prediction and ground-truth depth. However, per-pixel gradient loss does not always provide the expected benefits. For instance, DepthAnything v2\cite{yang2024depthv2} found that gradient matching loss improves depth sharpness when trained on synthetic data but fails to bring improvement when trained on labeled real datasets. Similarly, we find that directly applying MiDaS-style\cite{midas} gradient matching loss to our training paradigm does not improve sharpness. We hypothesize this is due to the persistence of small misalignment between the input and output that dampen the fine-grained supervision signal for gradient loss. 

\paragraph{Patch Gradient loss.} To remedy this, we construct a gradient loss that measures structural consistency across multiple spatial scales while remaining robust to local translation and scale differences (Algorithm~\ref{alg:gradloss}). Following from Depth Pro \cite{bochkovskii2025depthprosharpmonocular}, prediction and target maps are pre-processed using a log-space mean-centering transformation to eliminate sensitivity to global scale or offset. Spatial derivatives are then computed using horizontal, vertical, and diagonal Sobel operators. At the pixel level, we compute the absolute difference between predicted and target gradients. To capture higher-order spatial context and reduce sensitivity to small misalignment, we apply a differentiable range pooling operator that estimates a min-max range of Sobel gradients within $3\times3$ and $5\times5$ patches using softmax with a temperature hyperparameter. 
The complete loss combines pixel-wise gradient magnitude, patch-wise gradient range loss, and a unit-direction consistency to align the gradient orientations between prediction and target. The losses are computed at multiple image resolutions (1.0$\times, 0.5\times, 0.25\times$) similar to MiDaS-style gradient loss\cite{midas}.

\paragraph{Final loss.} The final loss is the combination of SiLog loss, progressively annealed and replaced by a Charbonnier loss, with the progressive addition of the Patch Gradient loss at mid training. The parameters of this curriculum loss, that we denote \ourloss, are provided in the next section.

\subsection{Decoder architecture, training details and data sampling}

\label{sec:architecture_training_changes}
The DINOv3 paper experimented with the same decoder as CHMv1 \cite{tolan2024very}, modified with a larger image resolution of $448\times448$ instead of $256\times256$, resulting in a large boost in the performance of all models (from 0.6 to 0.8 block $R^2$) on SatLidar v1. 
When introducing our new loss function to this setting, we encountered instability issues that motivated some of the architecture changes described below. 

\paragraph{Architecture changes.}
We bring slight architecture changes to the decoder compared to CHMv1. The most important change is the modification of the binning strategy from a linear scale to a mix of linear and log scale. Compared to CHMv1, where the CHMs ground truths were divided by 10, we use a factor of 8. We use a maximum depth of 96m/8 for the decoder predictions instead of 80m/10. We use bias in the residual layers. In the up-sampling head of the decoder (UpConvHead), we use Kaiming initialization. We remove the extra $1\times1$ convolution projection layer of the decoder. We increase the dimension of the final hidden layer of the UpConvHead from 32 to 128. We use intermediate backbone layers [5, 11, 17, 23] instead of [4, 11, 17, 23]. The backbone norm is set to true following the DINOv3 defaults.


\paragraph{Training details.}

We use a similar training loop to DINOv3 depth training \cite{dinov3code}, with a learning rate of $10^{-4}$, $\beta_1=0.9$, $\beta2=0.99$, weight decay of 0.001, 100k iterations, a total batch size of 16 (2 with 8 GPU), and cosine LR scheduler with 6k linear warmup iterations. 
In the curriculum loss, the SiLog loss is progressively replaced by Charbonnier loss during the first 30k iterations. The Patch Gradient loss is linearly warmed up with a weight from 0 to 0.075 between iterations 5k and 50k.  

\paragraph{Data sampling.}

We sample data from the concatenation of our curated NAIP-3DEP and SatLidar v2 datasets. We also implement batch category sampling, to ensure each batch contains some proportion of small and tall trees. The category sampling ratios were determined based on an ablation study [10$\%$ of GT below 1m, $20\%$ above 35m] for SatLidar and [10$\%$ of GT below 1m, $10\%$ above 25m] for NAIP-3DEP.

\begin{figure}[htb]
     \centering
     \begin{subfigure}[b]{0.48\textwidth}
     \caption{CHMv1 on NAIP-3DEP dataset}
         \centering
         \def\big{\includegraphics{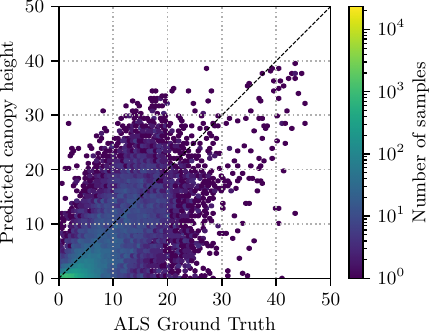}}
        \def\little{\includegraphics{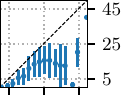}}
        \stackinset{l}{30pt}{t}{5pt}{\little}{\big}
     \end{subfigure}
     \hfill
     \begin{subfigure}[b]{0.48\textwidth}
         \centering
        \caption{CHMv2 on NAIP-3DEP dataset}
         \def\big{\includegraphics{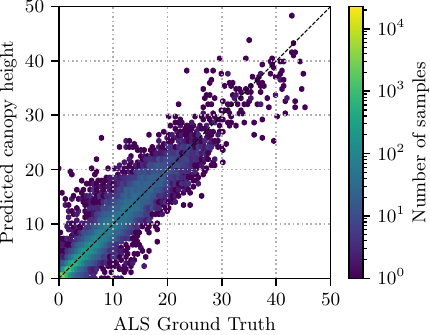}}
        \def\little{\includegraphics{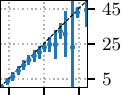}}
        \stackinset{l}{30pt}{t}{5pt}{\little}{\big}
     \end{subfigure}\\
     ~\\
     \begin{subfigure}[b]{0.48\textwidth}
     \caption{CHMv1 on SatLidar v2}
         \centering
         \def\big{\includegraphics{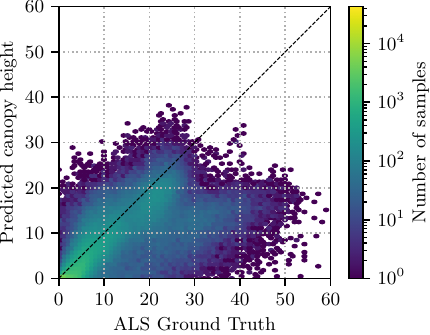}}
        \def\little{\includegraphics{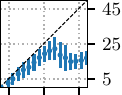}}
        \stackinset{l}{30pt}{t}{5pt}{\little}{\big}
     \end{subfigure}
     \hfill
     \begin{subfigure}[b]{0.48\textwidth}
         \centering
        \caption{CHMv2 on SatLidar v2}
         \def\big{\includegraphics{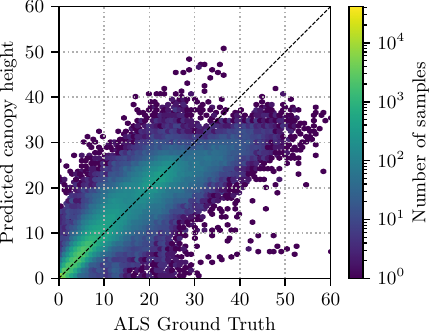}}
        \def\little{\includegraphics{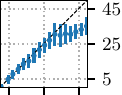}}
        \stackinset{l}{30pt}{t}{5pt}{\little}{\big}
     \end{subfigure}
      \caption{Compared to CHMv1 on NAIP-3DEP (top) and SatLidar v2 (bottom), CHMv2 exhibits greatly reduced biases, especially for trees above 30 m. The inset figures plot the 95th percentiles of CHM on $50\times50$ crops ($y$ axis) as a function of the 95th percentiles of Ground Truths ($x$ axis). Scales are in meters.}
     \label{fig:plots_3dep_satlidarv2}
\end{figure}

\section{Data Record} 
The CHMv2 dataset covers nearly the entirety of global land area (except Greenland and Antarctica) with canopy height values encoded in integer meters for each pixel. The 22.65 terabyte dataset is released in 213,109 cloud-optimized geotiffs (COGs) that are 32,768 $\times$ 32,768 pixel in size, with a 1.2m pixel spacing at the equator. Each COG covers approximately 1,500 km\textsuperscript{2} at the equator and about 65 km\textsuperscript{2} near the poles. Pixels where the input imagery was occluded by clouds, as specified by the Maxar Vivid metadata, are encoded by the no data value. Each file is released in a Pseudo-Mercator projection (EPSG 3857).

The CHMv2 dataset is available via Amazon Web Services (AWS)\footnote{\url{s3://dataforgood-fb-data/forests/v2/global/dinov3_global_chm_v2_ml3/chm/}} under the DINOv3 license.\footnote{\url{https://github.com/facebookresearch/dinov3/blob/main/LICENSE.md}} It is also released on a Google Earth Engine viewer \footnote{\url{https://meta-forest-monitoring-okw37.projects.earthengine.app/view/canopyheight}}. A global lookup file listing tile names and bounding boxes is included in the AWS repository. We additionally release a global GeoTIFF of input image acquisition date, where pixel values encode year minus 2000 (e.g., 18.25 indicates April 2018).

\section{Technical Validation}

\subsection{Improvements versus CHMv1}

\begin{figure}
\centering
\begin{tabular}{wc{3.5cm} wc{3.6cm} wc{3.6cm} wc{3.6cm} wc{1.8cm}}
Image & CHMv1 & CHMv2 & ALS Ground Truth& Meters
\end{tabular}\\
    \includegraphics[width=1\linewidth]{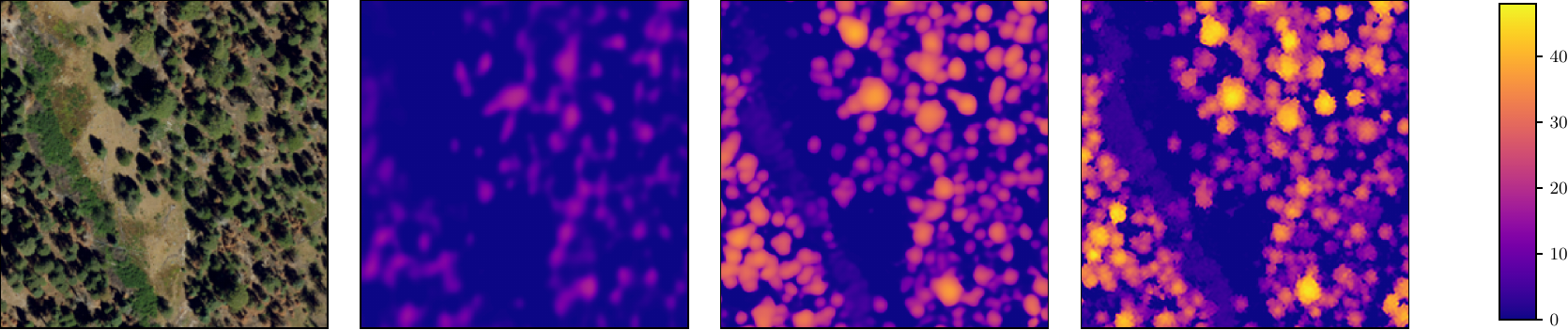}\\
    \vspace{0.2ex}
    \includegraphics[width=1\linewidth]{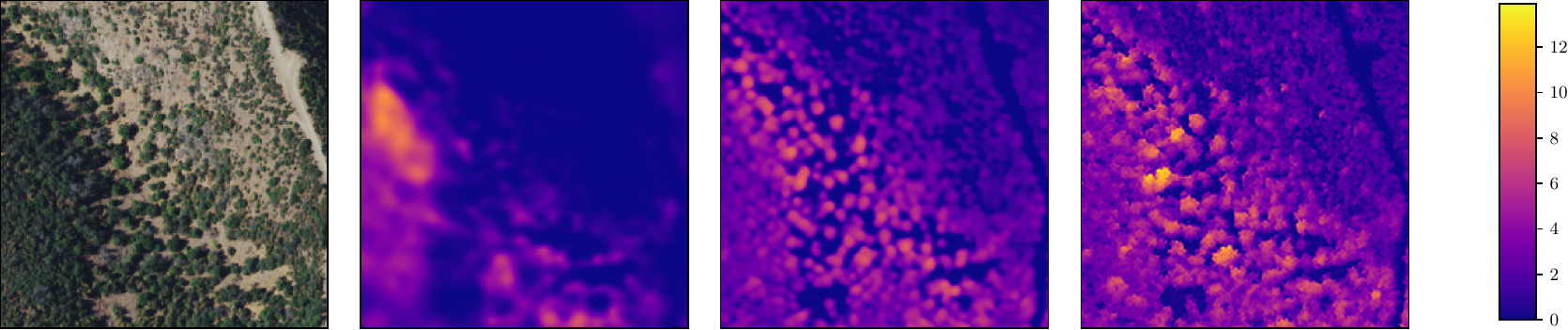}\\
    \vspace{0.2ex}
    \includegraphics[width=1\linewidth]{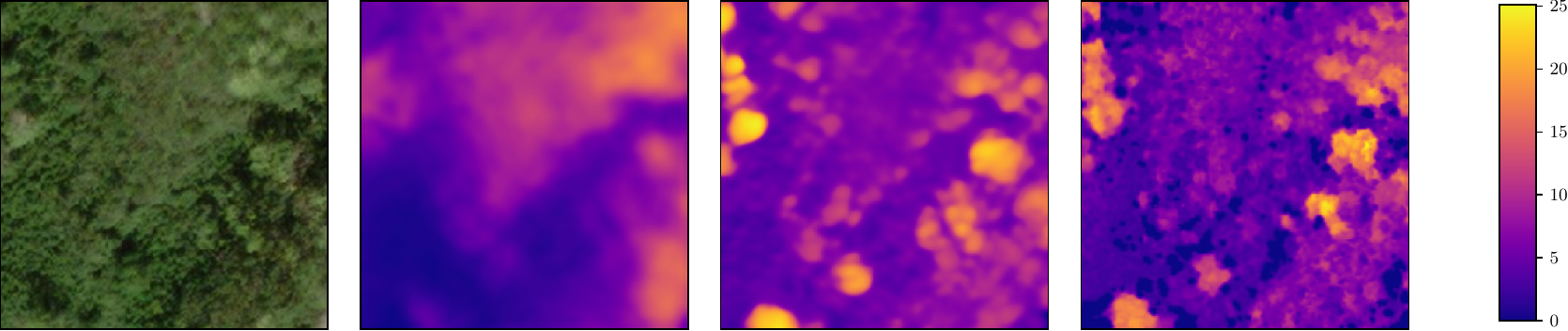}\\
    \vspace{0.2ex}
    \includegraphics[width=1\linewidth]{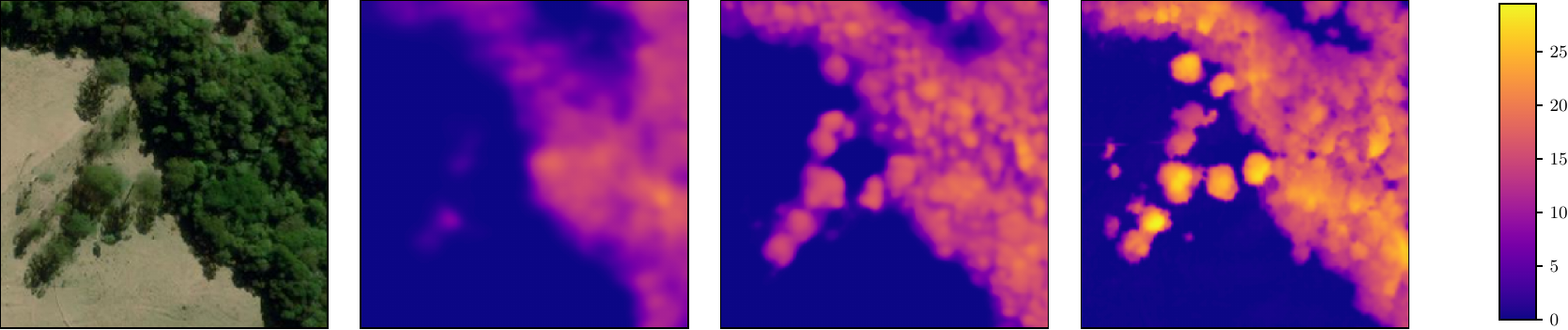}\\
    \caption{Qualitative improvements over CHMv1, in terms of sharpness and accuracy. Comparison on the NAIP-3DEP (two top lines) and SatLidar v2 (bottom lines) on 256$\times$256 samples. CHMv1 is without GEDI correction.}
    \label{fig:qualitative_comparison}
\end{figure}

\begin{figure}
    \centering
   \begin{tabular}{wc{3.5cm} wc{3.5cm} wc{3.5cm} wc{3.5cm} wc{1.7cm}}
Satellite view 1 & Satellite view 2 & Aerial view & ALS Ground Truth& Meters
\end{tabular}
    \includegraphics[width=0.97\textwidth]{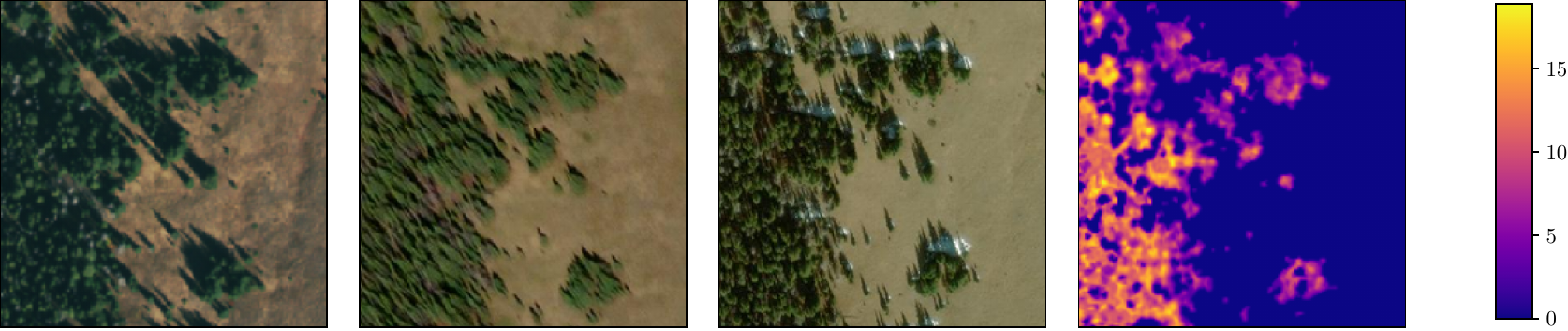}\\
    ~\\
    \begin{tabular}{cc}
    CHMv1 from view 1, view 2 and aerial images &
    CHMv2 from view 1, view 2 and aerial images\\
    \includegraphics[width=0.485\textwidth]{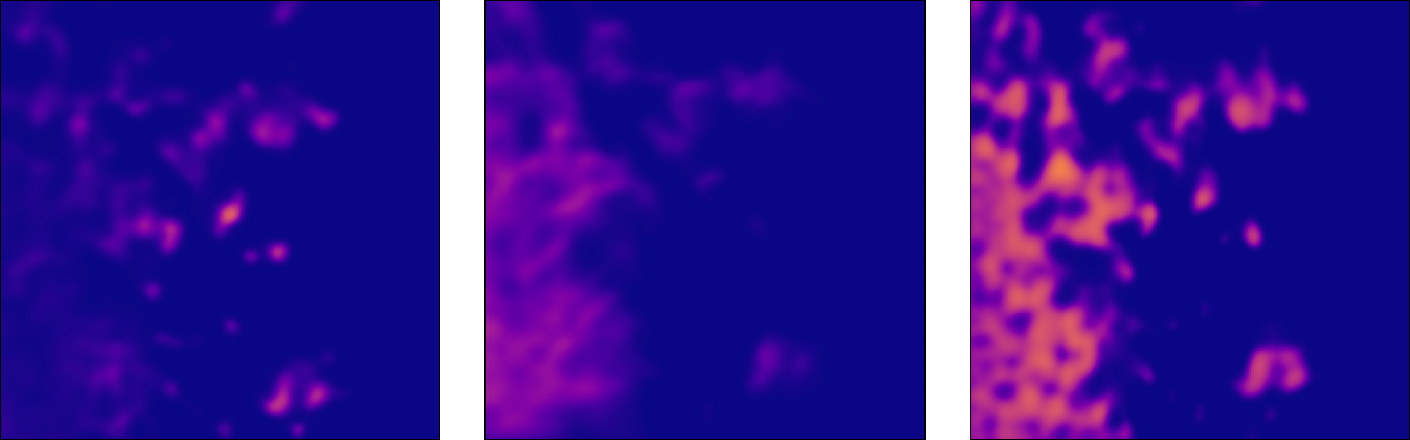}&
    \includegraphics[width=0.485\textwidth]{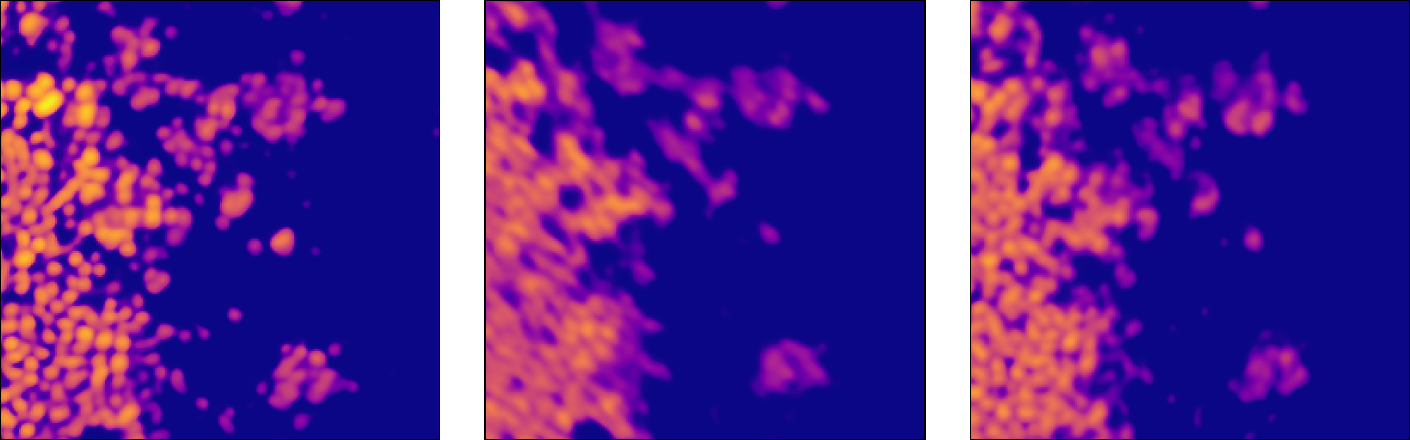}\\
     \hspace*{-.07\textwidth} \begin{tabular}{c}
         \begin{minipage}{0.15\textwidth}
             \begin{tabular}{c}
            \def\big{\includegraphics[width=\textwidth]{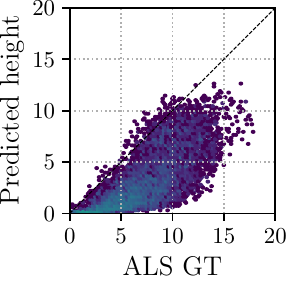}}
            \def\little{\includegraphics[width=0.32\textwidth]{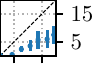}}
            \stackinset{l}{20pt}{t}{4pt}{\little}{\big}
            \end{tabular}
         \end{minipage}
         \begin{minipage}{0.15\textwidth}
             \begin{tabular}{c}
             \def\big{\includegraphics[width=\textwidth]{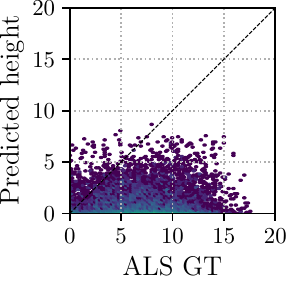}}
            \def\little{\includegraphics[width=0.32\textwidth]{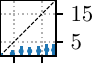}}
            \stackinset{l}{20pt}{t}{4pt}{\little}{\big}
            \end{tabular}
         \end{minipage}
         \begin{minipage}{0.15\textwidth}
             \begin{tabular}{c}
              \def\big{\includegraphics[width=\textwidth]{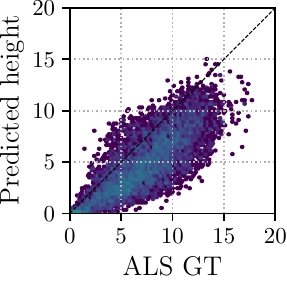}}
            \def\little{\includegraphics[width=0.32\textwidth]{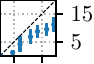}}
            \stackinset{l}{20pt}{t}{4pt}{\little}{\big}
            \end{tabular}
         \end{minipage}
     \end{tabular}
     & \hspace*{-.08\textwidth}
     \begin{tabular}{c}
        \begin{minipage}{.15\textwidth}
        \begin{tabular}{c}
        \def\big{\includegraphics[width=\textwidth]{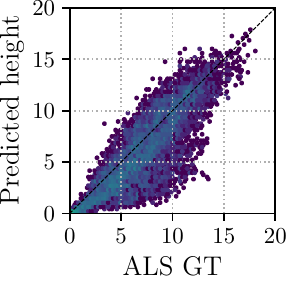}}
        \def\little{\includegraphics[width=0.32\textwidth]{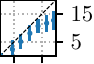}}
        \stackinset{l}{20pt}{t}{4pt}{\little}{\big}
        \end{tabular}
     \end{minipage}
     \begin{minipage}{0.15\textwidth}
         \begin{tabular}{c}
         \def\big{\includegraphics[width=\textwidth]{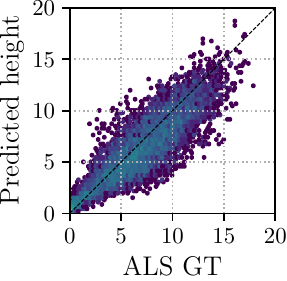}}
        \def\little{\includegraphics[width=0.32\textwidth]{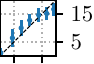}}
        \stackinset{l}{20pt}{t}{4pt}{\little}{\big}
        \end{tabular}
     \end{minipage} 
        \begin{minipage}{0.15\textwidth}
         \begin{tabular}{cc}
         \def\big{\includegraphics[width=\textwidth]{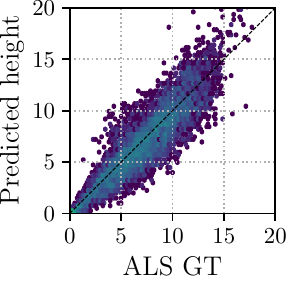}}
        \def\little{\includegraphics[width=0.32\textwidth]{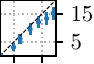}}
        \stackinset{l}{20pt}{t}{4pt}{\little}{\big}&
        \hspace{-1em}
         \includegraphics[width=0.27\textwidth]{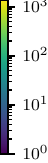}  
         \end{tabular}
      \end{minipage}
      \end{tabular}
    \end{tabular}
    \caption{Comparison of CHMv1 and CHMv2 on the Multisource dataset. CHMv1 is without GEDI correction.}
     \label{fig:plots_multisource}
\end{figure}

 The presented dataset has large overall performance gains relative to CHMv1\cite{tolan2024very} measured against our ALS-derived test datasets. On the SatLidar test dataset, the MAE improves from 4.3m to 3.0m, with $R^2$ improvements from 0.53 to 0.86 (Table~\ref{tab:main_ablation}, Figure~\ref{fig:plots_3dep_satlidarv2}) compared to CHMv1. The CHMv2 model and data also has a marked improvement in mean bias error at high ($\geq$ 30m) height (Figure~\ref{fig:plots_3dep_satlidarv2}).  We still note area for improvement in very high ($\ge$ p98) canopy estimation, which are still underestimated. When qualitatively compared with CHMv1, the presented dataset has noticeable improvements in crown delineation, characterization of canopy gaps and complex canopy structures, and improved sharpness of canopy edges (Figure~\ref{fig:qualitative_comparison}). Finally, CHMv2 performs much better than CHMv1 on generating consistent CHM predictions across multiple sources of imagery of the same location (Figure~\ref{fig:plots_multisource}).  

\subsection{Comparison with other global datasets}

\begin{figure}[!htbp]
    \begin{center}
    \includegraphics[width=.9\linewidth]
    {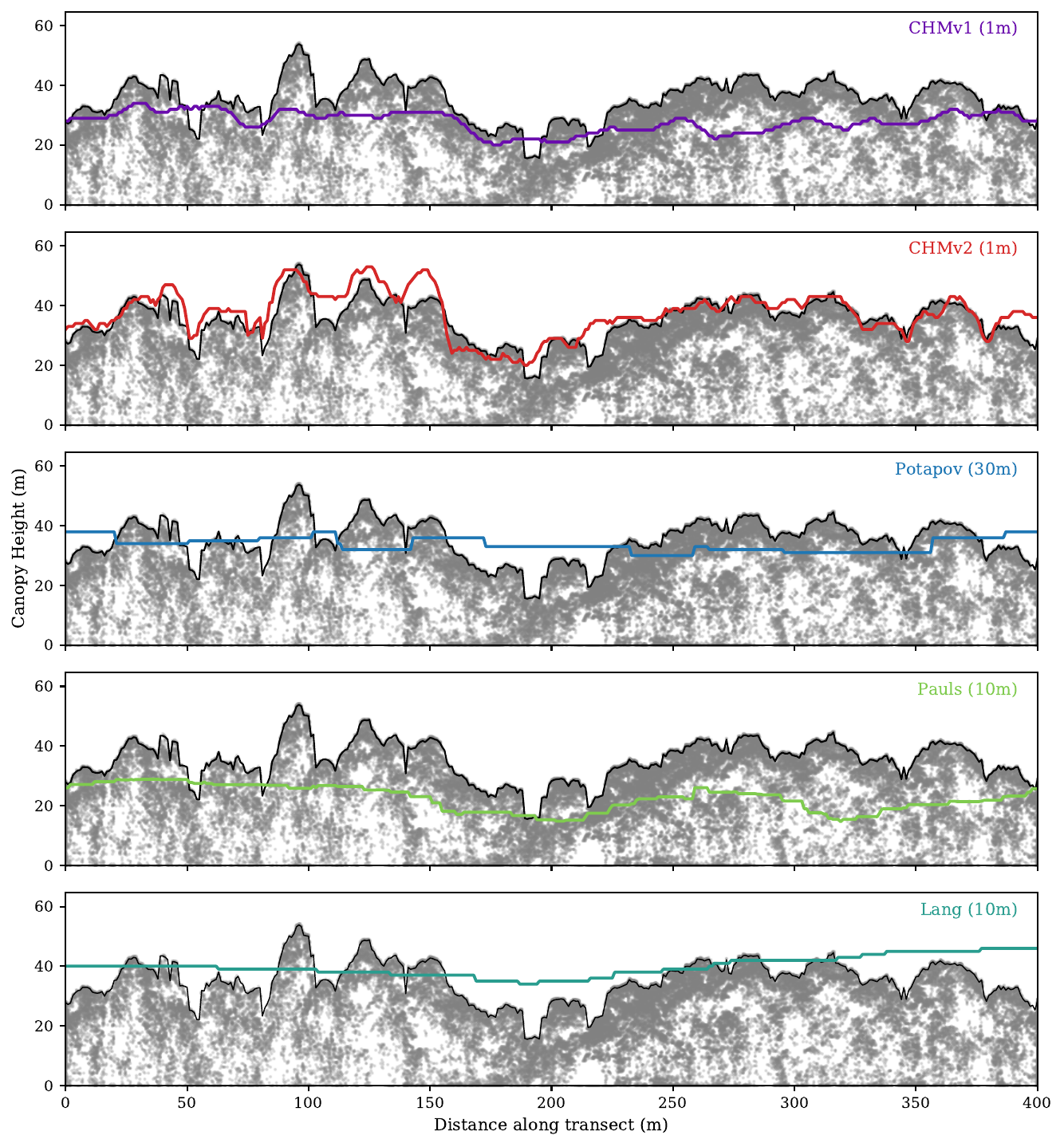}
    \end{center}
    \caption{
       Canopy height profiles along a single 10-m wide transect comparing ALS reference measurements to five canopy height datasets. Gray points show ALS returns within the transect corridor; the black line shows the ALS-derived canopy height profile. ALS data is from Kalimantan, Indonesia (1.6412$\degree$, 115.281$\degree$) and is $\ge 200$ km from any observation within the SatLidar v2 dataset.
    }
    \label{fig:transect}
\end{figure}

We compare CHMv2 against existing low-resolution global canopy height products, including Potapov \textit{et al.} \cite{Potapov2021Mapping}, Paul \textit{et al.} \cite{pauls2024estimatingcanopyheightscale}, and Lang \textit{et al.} \cite{Lang2022High}, by evaluating all products on the SatLidar v2 ALS-derived test set (Table~\ref{tab:comparison}). This provides a direct comparison between CHMv2 and prior global canopy height products on a shared meter-scale ALS reference. The compared low-resolution products are trained using spaceborne LiDAR supervision (e.g., GEDI) and are primarily evaluated in their original manuscripts against held-out GEDI footprints. In contrast, CHMv2 (and CHMv1) is trained solely on ALS-derived canopy height maps. We therefore evaluate all products against the same ALS-derived reference data to ensure a consistent comparison on our test distribution. CHMv2 achieves the lowest error on the SatLidar v2 ALS-derived test split (MAE = 3.0 m), improving over the evaluated low-resolution global products (4.9 - 8.4 m). A visual comparison of the evaluated datasets demonstrates the improved capability of CHMv2 for measuring forest structure when compared to ALS-derived canopy height (Figure \ref{fig:transect}).

\begin{table}[ht]
\centering
{
\begin{tabular}{@{}lcc@{}}
\toprule
\textbf{Dataset} & \textbf{Resolution (m)} &\textbf{MAE} \\
\midrule
Pauls {\it \textit{et al.}}~\cite{pauls2024estimatingcanopyheightscale} & 10 &7.5\\
Lang {\it \textit{et al.}}~\cite{Lang2022High} & 10 &8.4\\
Potapov {\it \textit{et al.}}~\cite{Potapov2021Mapping} & 30 &4.9\\
CHMv1~\cite{tolan2024very} &  1&4.3\\
CHMv2~ &  1&\textbf{3.0} \\
\bottomrule
\end{tabular}}
\caption{Comparison of global canopy height products to ground truth ALS-derived CHM data in the SatLidar v2 test split.}
\label{tab:comparison}
\end{table}

\subsection{Comparison with GEDI and ICESat-2}

To evaluate agreement between CHMv2 and canopy height from spaceborne LiDAR, we compiled reference observations from GEDI (2019–2022) and ICESat-2 (2018–2023). We used the footprint-level RH98 metric from the GEDI L2A product\cite{Dubayah2021GEDIL2A} and the 98th percentile canopy height for 20-m along-track segments from ICESat-2 ATLAS ATL08\_V7\cite{Neuenschwander2025-yi}. Only high-quality observations were retained. For GEDI, we selected full-power beam, nighttime, leaf-on shots with sensitivity $\ge$ 0.95, the highest quality flag, and a difference between the DEM and lowest detected elevation $<$ 150 m. For ICESat-2, we selected strong-beam nighttime segments and excluded observations flagged for snow/ice, aerosols, or clouds.

To ensure geographically balanced sampling, we randomly extracted 10,000 GEDI observations and 5,000 ICESat-2 observations per $1\degree \times 1\degree$ grid cell. For each reference observation, we computed the 98th percentile of CHMv2 within a 12 m radius (GEDI) or 7 m radius (ICESat-2). We excluded samples where tree cover loss or gain\cite{doi:10.1126/science.1244693, Potapov2021Mapping} indicated potential forest change between the Maxar acquisition and the ALS observation date. To reduce sensitivity to Maxar image quality and seasonality, we also excluded samples with cloud contamination, sun elevation $< 45\degree$, or off-nadir angle $>25\degree$. Additional exclusions removed samples where (i) the tree-cover edge intersected the footprint, or (ii) the reference canopy height was inconsistent with existing land cover and tree height products\cite{Potapov2021Mapping, Zanaga2022-pa}. These quality and consistency filters removed 66\% of GEDI samples and 78\% of ICESat-2 samples, reflecting strict screening for cloud contamination, acquisition geometry, and temporal mismatch between imagery and reference observations. For the remaining observations, reference height was set to zero within urban, bare ground, and water classes\cite{Zanaga2022-pa} to reduce contamination from non-woody vertical structures.

\begin{figure}[H]
\begin{center}
    \begin{subfigure}[b]{0.9\textwidth}
    \caption{
            Comparison of CHMv2 98$^{\mbox{th}}$ percentile with GEDI Relative Height (RH) at the 98$^{\mbox{th}}$ percentile.
        }
        \begin{center}
        \vspace{-1ex}
        \includegraphics[width=.95\linewidth]
        {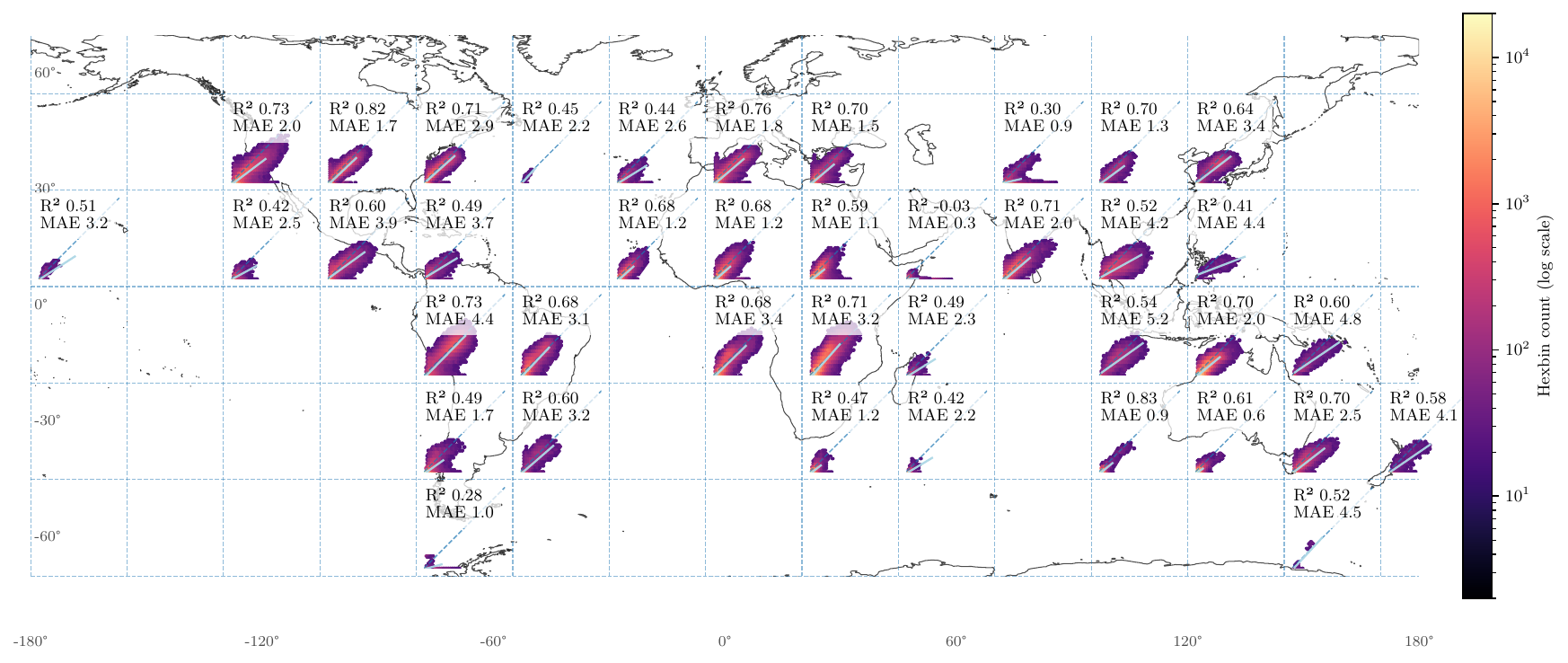}
        \end{center} 
    \end{subfigure}\\
    \begin{subfigure}[b]{0.9\textwidth}
    \caption{
            Comparison of CHMv2 98$^{\mbox{th}}$ percentile canopy height (y-axis) against ICESat-2 98$^{\mbox{th}}$ percentile relative height (x-axis).
        }
        \begin{center}
        \includegraphics[width=.95\linewidth]
        {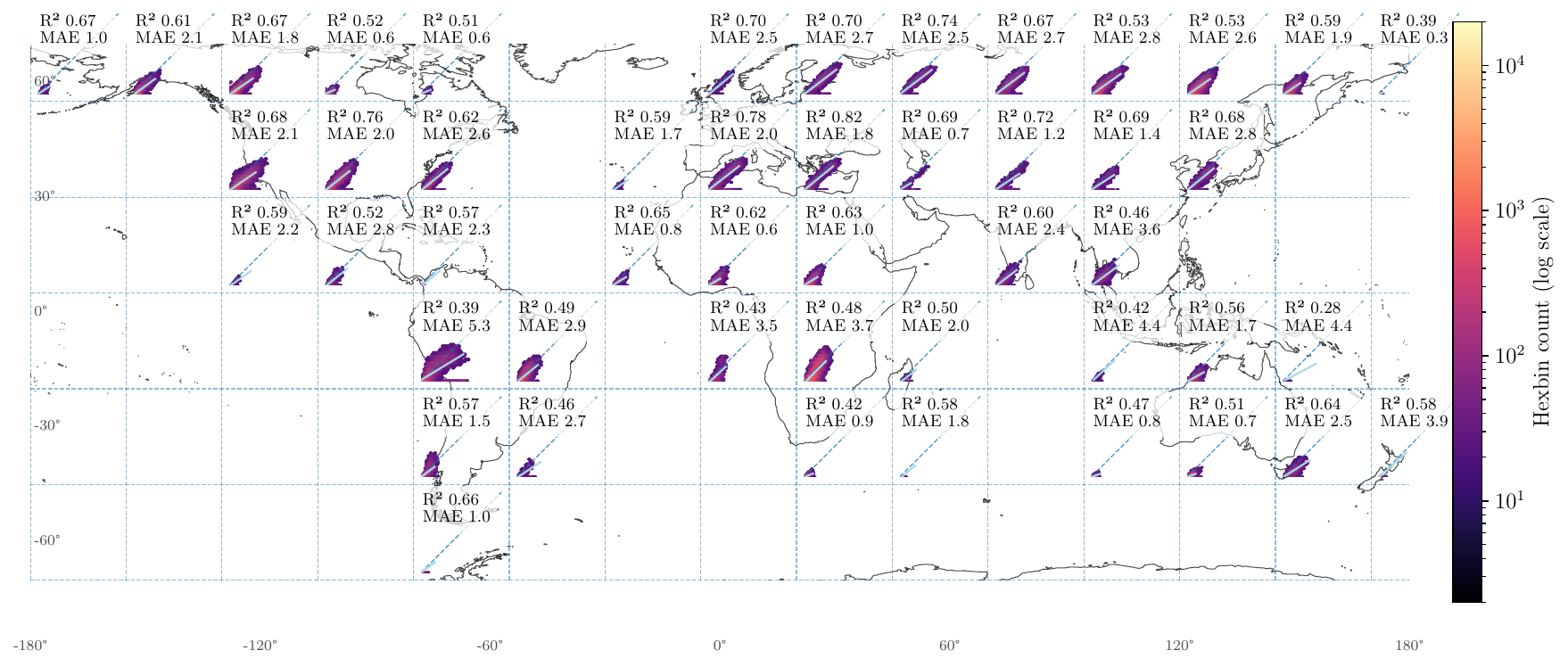}
        \end{center}
         
    \end{subfigure}
    \end{center}
    \caption{Comparison with GEDI and ICESat-2.}
    \label{fig:gedifigure}
\end{figure}

We evaluated agreement using 37M GEDI and 11M ICESat-2 reference samples. Global comparison against GEDI yielded an $R^2$ of 0.70, MAE of 3.1 m, and RMSE of 6.4 m. Regional comparisons (Figure \ref{fig:gedifigure}a) showed consistently high agreement across major forest biomes ($R^2$ = 0.68–0.76), with higher uncertainty in sparsely treed landscapes. Global comparison against ICESat-2 yielded similar agreement with an MAE of 2.9 m and RMSE of 5.63 m, but a lower $R^2$ of 0.60. The lower $R^2$ may partially be explained by the smaller footprint size of ICESat-2 introducing more random geolocation noise. Regional comparisons of ICESat-2 (Figure \ref{fig:gedifigure}b) show strong agreement in boreal and temperate regions with slightly lower agreement in the tropics, while the regional comparisons with GEDI show relatively stronger agreement in the tropics.

GEDI and ICESat-2 provide the only globally consistent reference canopy height observations, but their estimates are not strict ground truth. Retrieval accuracy is affected by slope, canopy density\cite{LIU2021112571}, and non-woody vertical structures\cite{Chen2025-qj}. Relative to airborne laser scanning, ICESat-2 generally exhibits higher canopy height uncertainty than GEDI\cite{LIU2021112571}, but it provides coverage in high-latitude regions not observed by GEDI. Both sensors are also affected by geolocation uncertainty: mean GEDI geolocation error is $\approx$10 m\cite{Tang2023-gl}, while ICESat-2 error is $\approx$2.5–4.4 m\cite{https://doi.org/10.1029/2020EA001494}. Given the mean Maxar geolocation accuracy of 8.7 m for this dataset, compound Maxar–lidar geolocation errors may substantially affect agreement within individual footprints.

\subsection{Ablation results}

Our ablation studies identify the relative contributions of training data quality, migration from DINOv2 to DINOv3, loss design, and training curriculum on our qualitative and quantitative results.  We employ the same metrics as Tolan \textit{et al.} \cite{tolan2024very} to measure ablation results, namely MAE, block-$R^2$, bias, and Edge Error, which measures the difference between the Sobel gradient of the GT and prediction.

After comparing results using a ViT-L and larger models in the DINOv3 study~\cite{simeoni2025dinov3} on SatLidar v1, we concluded that using larger backbones would lead to very similar results with significant compute overhead. Specifically, there was no performance improvement when using the CHMv1 ViT-H backbone instead of ViT-L. Therefore, we use only ViT-L backbones for CHMv1 and our approach in this section. We compared the different models without post-processing (no GEDI correction step~\cite{tolan2024very}).

\begin{table}[h]
\centering
\resizebox{\columnwidth}{!}
{
\begin{tabular}{@{}lcccc  c cccc c ccc@{}}
\toprule
 \multirow{2}[3]{*}{\textbf{Backbone}} & 
\multirow{2}[3]{*}{{\bf Setting}} &
\multirow{2}[3]{*}{\textbf{Loss}} & 
\multirow{2}[3]{*}{\textbf{Sampling}} &
\multirow{2}[3]{*}{\textbf{Train data}} &&
\multicolumn{4}{c}{\textbf{SatLidar v2}} && \multicolumn{3}{c}{\textbf{NAIP-3DEP}} \\
 \cmidrule{7-10}  \cmidrule{12-14}
 &   &&  &   && {MAE} & \textbf{$R^2$}& Bias & Edge && {MAE} & \textbf{$R^2$} & Edge\\
\midrule
CHMv1 & Base &  SiLog & Uniform & Neon train && 4.3 & 0.53 &2.6& 0.67 && 2.2 & 0.54 & 0.77\\ 
DINOv3 & Base & SiLog & Uniform & Neon train && 3.9 & 0.64 &2.2&0.66&& 2.0 & 0.59 & 0.72 \\ 
DINOv3 & Base  & SiLog & Uniform & SatLidar v1 train && 3.5 & 0.78 &0.9&0.62 && 2.0 & 0.46 & 0.63 \\ 
DINOv3 & Base  & SiLog & Uniform & SatLidar v2 train && 3.5 & 0.76 &0.5&0.58 && 2.1 & 0.52 & 0.63\\ 
DINOv3 &Final &  \ourloss~ & Uniform & SatLidar v2 train && 3.5 & 0.75 &-0.5&0.55 && 2.2 & 0.48& 0.58 \\ 
DINOv3 &Final & \ourloss~ & Cat & 3DEP, SatLidar v2 train && 3.1 & 0.85 &0.1& 0.55 && 1.4 & {\bf 0.94}& {\bf 0.53}\\ 
DINOv3 &Final &  \ourloss~ & Cat & 3DEP, SatLidar v2 trainval && {\bf 3.0} & {\bf 0.86} &{\bf0.0} &{\bf 0.53}&& {\bf 1.4} & 0.93 & {\bf 0.53}\\ 
\bottomrule
\end{tabular}}
\caption{Main ablation from the CHMv1 settings to the CHMv2 settings, demonstrating improvements due to the DINOv3 backbone, training data diversity, registration, new architecture and loss. The base setting corresponds to the CHMv1 architecture and training parameters, and the final setting to the changes described in Section~\ref{sec:architecture_training_changes}.}
\label{tab:main_ablation}
\end{table}

Table~\ref{tab:main_ablation} summarizes the improvements from CHMv1 using DINOv2. The first two lines compare models trained on Neon, demonstrating improvements brought by training the DINOv3 backbone in all metrics, e.g. from 0.53 to 0.64 of $R^2$, and significantly sharper results on NAIP-3DEP.     
Second, the replacement of the Neon data by the more diverse SatLidar v1 data results in another large boost of $R^2$ on SatLidar v2 (0.64 to 0.78); however, we also note a drop of overall accuracy on the out of domain performance on NAIP-3DEP, except an improved edge error.
Third, the registration and quality filtering of the SatLidar dataset leads to a reduction of the edge error on SatLidar v2 and improvement in $R^2$ on NAIP-3DEP. Changes in architecture, training setting, and loss function bring only improvements in terms of edge error. The combination of aerial and satellite datasets for training leads to drastic improvements in all metrics (e.g. from 0.48 to 0.94 of $R^2$ on NAIP-3DEP, and from 0.75 to 0.85 on SatLidar v2). Our final model was trained on this data mix, augmented by the NAIP Sea dataset described above, and the val set of the SatLidar v2 dataset, leading to a small boost on SatLidar v2.

\begin{figure}[ht]
    \begin{center}
     \begin{minipage}{0.22\textwidth}
     \begin{center}
     Image
     \end{center}
    \end{minipage}%
    \begin{minipage}{0.22\textwidth}
    \begin{center}
   SiLog loss \\
   Edge error: 0.536
   \end{center}
    \end{minipage}
    \begin{minipage}{0.22\textwidth}
    \begin{center}
    Our loss\\
    Edge error: 0.527
    \end{center}
    \end{minipage}%
    \begin{minipage}{0.22\textwidth}
    \begin{center}
    ALS Ground Truth\\
    \end{center}
    \end{minipage}
    \begin{minipage}{0.1\textwidth}
    \begin{center}
    ~\\
    Meters\\
    \end{center}
    \end{minipage}
    \includegraphics[height=0.2\textwidth]{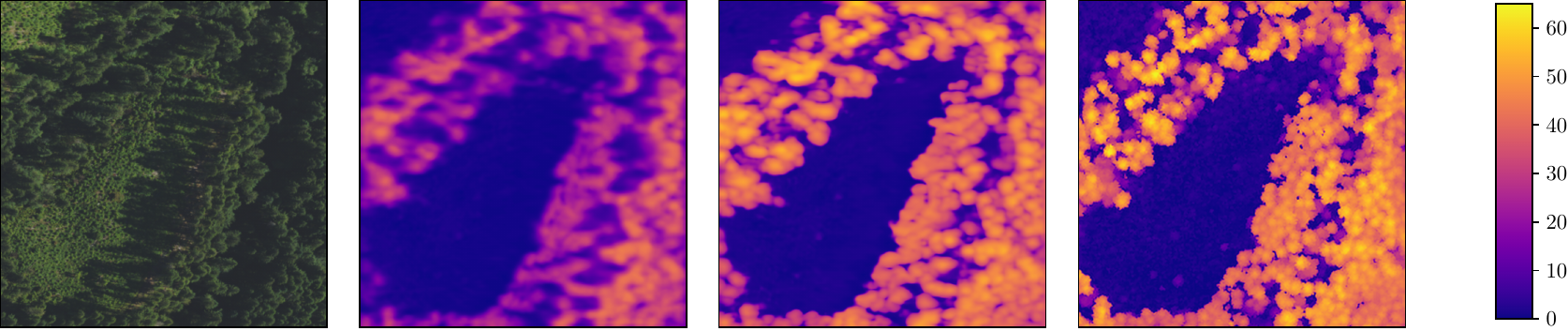}\\
    \includegraphics[height=0.2\textwidth]{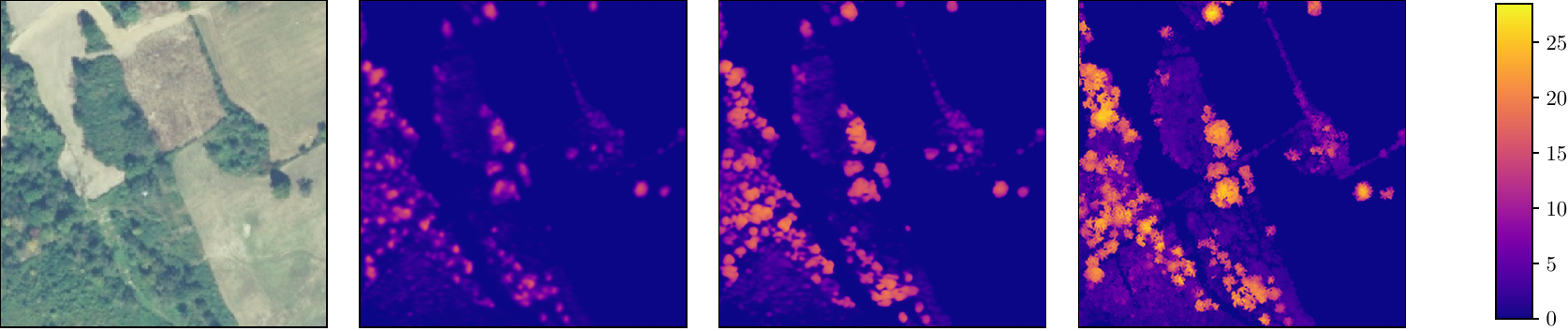}\\
    \end{center}
    \caption{Impact of our loss, a Combination of SiLog, Charbonnier and Multi-scale Patch Gradient on the NAIP-3DEP results. Compared to SiLog loss, results are sharper and more accurate.}
    \label{fig:lossexample}
\end{figure}

Table~\ref{tab:backbone_comparison} details that  training decoders with larger image resolution leads to important gains, e.g. $R^2$ from 0.73 to 0.81. We also note the large improvement of bias using the DINOv3 backbone on the out of domain NAIP-3DEP aerial dataset.

\begin{table}[ht]
\begin{minipage}{.5\textwidth}
\resizebox{0.95\columnwidth}{!}
{
\begin{tabular}{@{}l c c ccc c ccc@{}}
\toprule
\multirow{2}[3]{*}{{\bfseries\makecell{Back\\bone}}} & \multirow{2}[3]{*}{\bfseries\makecell{Decoder\\data}} && \multicolumn{3}{c}{\textbf{SatLidar v1}} && \multicolumn{3}{c}{\textbf{NAIP-3DEP}}  \\
\cmidrule{4-6} \cmidrule{8-10}
  &   &    &{MAE} & \textbf{$R^2$}& Bias && {MAE} & \textbf{$R^2$} & Bias\\
\midrule
CHMv1 &   Neon 256$\times$&& 4.0 & 0.61 & 3.7 && 2.21 & 0.54 & 1.1\\ 
CHMv1  & Sat v1  256$\times$ && 3.6 & 0.73 & 1.2 && 2.21 & {\bf 0.59} & 0.5\\ 
CHMv1  & Sat v1 448$\times$ && 3.4 & {\bf 0.81} & 0.9 && 2.32 & 0.52 & 0.9\\ 
DINOv3  & Sat v1 448$\times$ && \bf 3.2 & \bf 0.81 & {\bf 0.8} && {\bf 2.17} & 0.52 & {\bf 0.0}\\ 
\bottomrule
\end{tabular}}
\caption{
(a) Backbone ablation. Factors of improvements: {\bf more diverse training data, larger decoder training resolution, improved DINOv3 backbone}. 
Loss: SiLog.} 
\label{tab:backbone_comparison}
\end{minipage}%
\hfill
\begin{minipage}{0.5\textwidth}
\resizebox{0.95\columnwidth}{!}
{
\begin{tabular}{@{}lccccccc@{}}
\toprule
\multirow{2}[3]{*}{{\bfseries\makecell{Back\\bone}}} & \multirow{2}[3]{*}{{\bfseries{Setting}}} & \multirow{2}[3]{*}{\bfseries\makecell{Loss}}  && \multicolumn{4}{c}{\textbf{NAIP-3DEP}} \\
\cmidrule{5-8}
  & &   && MAE & \textbf{$R^2$} & Bias & Edge\\
\midrule
DINOv3  & Base &  SiLog && 1.61 & 0.82 & 1.0 & 0.55\\ 
DINOv3   & 100k iter & SiLog && 1.55 & 0.84 & 0.9 & 0.54\\ 
DINOv3  &  Final  & SiLog && 1.55 & 0.84 & 0.9 & 0.54\\ 
DINOv3   & Final  & \ourloss~ && {\bf 1.38} & {\bf 0.94} & {\bf 0.1} & {\bf 0.53}\\
\bottomrule
\end{tabular}}
\caption{(b) Decoder parameters ablation with decoders trained on the NAIP-3DEP dataset with 416$\times$ samples. Factors of improvement: {\bf longer training and loss.}}
\label{tab:training_ablation}
\end{minipage}
\end{table}

Table~\ref{tab:training_ablation} shows that moving from 38K to 100K training iterations is helpful. The architecture changes do not affect performance on this dataset; however, the loss brings a large boost in all metrics.   
As shown in Table~\ref{tab:architecture_ablation}, the architecture changes help or do not degrade performance on SatLidar v2. Table~\ref{tab:loss_ablation} explores different loss alternatives. Using a SiLog loss alone results in a large bias and relatively low $R^2$. Adding a gradient loss in an attempt to obtain sharper results mildly improves results but leads to larger edge error. Combining SiLog and Charbonnier greatly reduces the bias from 0.9 to 0.1; however results are still blurry as shown by the large 0.54 edge error. Adding a gradient loss term does not solve the problem, and neither does a patch gradient loss that creates grid artifacts. Our curriculum loss combines SiLog, Charbonnier and only starts enforcing a Patch Gradient loss at mid training, avoiding artifacts while yielding  sharper predictions (Figure~\ref{fig:lossexample}). Finally, Table~\ref{tab:mix_ablation} studies the importance of using category sampling and different aerial / satellite data ratios used in decoder training.

\begin{table}[ht]
\begin{minipage}{.415\textwidth}
\resizebox{1\columnwidth}{!}{
    \begin{tabular}{@{}lcccc@{}}
    \toprule
 \multirow{2}{*}{{\bfseries{Model}}} && \multicolumn{3}{c}{\textbf{SatLidar v2}}\\
  \cmidrule{3-5} 
         	&& MAE	 & Bias	& $R^2$ \\
            \midrule
Final Decoder arch. : setting F  &&	{\bf 3.15}	& -0.3 &{\bf 0.84}\\
\midrule
F + project	&&{\bf 3.15}	&  -0.3 &	{\bf 0.84}\\
F - bias in residual layers &&	3.16 &	{\bf-0.0} &	0.83\\
F - K. init in UpConvHead	&&3.20	& -0.1	& 0.82\\
F - larger dim of UpConvHead && 3.16 & {\bf-0.0} & 0.83 \\
F - backbone norm &&  3.16 & -0.1 &0.83\\
F - mixed bin mixing && 3.28 &{\bf-0.0}& 0.80\\
\bottomrule
    \end{tabular}}
     \caption{Most architecture changes do not affect performance except  bin mixing and Kaiming initialization.  Models trained on SatLidar~v2, 3DEP mix with Charbonnier and Gradient loss.}
     \label{tab:architecture_ablation}
\end{minipage}
\hfill
\begin{minipage}{.555\textwidth}
\centering
\resizebox{1\columnwidth}{!}
{
\begin{tabular}{@{}lcccccc@{}}
    \toprule
    \multirow{2}[3]{*}{\bfseries\makecell{Loss}}  && \multicolumn{5}{c}{\textbf{NAIP-3DEP}} \bigstrut \\
    \cmidrule{3-7}
      &&	MAE &	Bias &	$R^2$ & Edge & No Artifacts\\
        \midrule
        SiLog (S) && 1.55	&0.9	&0.84&	0.536 & \cmark\\ 
S + Grad &&		1.51&	0.8	&0.85 & 0.557 & \cmark\\
S + Char &&		1.35&	{\bf0.1}	&{\bf0.94} & 0.541 & \cmark\\
S + Char + Grad &&	{\bf 1.31}&	0.2	&{\bf 0.94} & 0.542& \cmark\\
S + Char + Patch Grad && 1.42&	{\bf 0.1}	&{\bf 0.94} & 0.507& \xmark\\
\midrule
\ourloss S + Char + Patch Grad&&		1.38&	{\bf0.1}	& {\bf0.94} & {\bf 0.527}& \cmark\\ 
\bottomrule
    \end{tabular}}
    \caption{Loss ablation on the NAIP-3DEP dataset. The combination of SiLog, Charbonnier and Patch gradient loss is the best compromise between accuracy and sharpness but suffers from artifacts, removed by the curriculum learning.}
    \label{tab:loss_ablation}
\end{minipage}
\end{table}

\begin{table}[htb]
    \centering
\begin{tabular}{@{}c c c   c   cc   c   cc  c  cc@{}}
\toprule
  \multirow{2}[3]{*}{{\bfseries\makecell{Category sampling}}} & \multirow{2}[3]{*}{\textbf{NAIP-3DEP}} &  \multirow{2}[3]{*}{{\bfseries\makecell{SatLidar v2}}} && \multicolumn{2}{c}{\textbf{SatLidar v2}}  && \multicolumn{2}{c}{\textbf{NAIP-3DEP}} && \multicolumn{2}{c}{\textbf{Multisource}}\\
  \cmidrule{5-6} \cmidrule{8-9}  \cmidrule{11-12}
    &   &  && MAE & $R^2$ && MAE & $R^2$  && MAE & $R^2$\\
\midrule
\xmark & 50$\%$ & 50$\%$ && {\bf 3.07} & 0.83       && 1.37 & {\bf 0.93} && 2.98  & 0.77  \\ 
\cmark & 50$\%$ & 50$\%$ && {\bf 3.07} & {\bf 0.84}   && 1.36 & {\bf 0.93} && 2.98 & 0.78 \\ 
\cmark & 30$\%$ & 70$\%$ && 3.10 & 0.83             && 1.38 &  {\bf 0.93} && {\bf 2.94} & {\bf 0.82} \\ 
\cmark & 70$\%$ & 30$\%$ && 3.18 & 0.82             && {\bf 1.35} & {\bf 0.93} && 2.98 & 0.79\\ 
\bottomrule
\end{tabular}
\caption{ Data mix ablation: using a 50-50 ratio of data source brings a good compromise. Category sampling shows slightly improved metrics. To complement this ablation, we computed that the category sampling improves the MAE of trees above 35 meters by at least two meters. Using DINOv3 viT-L backbone, Charbonnier + Gradient Loss.}
    \label{tab:mix_ablation}
\end{table}



\section{Usage Notes}

CHMv2 can be used either as a global meter-scale canopy height product, or as a pretrained model that can be applied to user-provided high-resolution imagery. Because CHMv2 is derived from single-date optical imagery, users should account for variability in image acquisition date, viewing geometry, and atmospheric condition. Although we apply cloud mask filtering, localized artifacts (e.g., haze, missed clouds, or seamlines between neighboring acquisitions) may persist in some regions.

When using CHMv2 to compute canopy height statistics, we recommend masking non-vegetated areas such as open water, built-up areas, or bare ground using an independent land cover map. CHMv2 can support a range of downstream analyses, including tree cover and land cover mapping, forest type segmentation, carbon and biomass estimation, assessments of forest structural diversity, and monitoring of tree-based land uses such as plantations and agroforestry systems. 

In addition to the canopy height product, we release the trained CHMv2 model to enable inference on user-provided satellite or aerial imagery at approximately 0.6\,m ground sampling distance (GSD). For applications requiring temporal change detection, users should preferentially compare imagery acquired under similar seasonal and illumination conditions.

\subsection{Example applications}
CHMv2 provides a complementary structural signal that can improve land use and land cover characterization when combined with spectral and texture features. For example, in agroforestry systems, height heterogeneity and multi-strata canopy structure can serve as indicators of shade tree presence and management intensity. Similarly, canopy height metrics can provide useful proxies for forest successional stage or stand development, supporting classification of secondary forest and inference about regrowth trajectories \cite{moudry,tian2023}. More broadly, CHMv2 can be used to derive spatial covariates for biomass estimation, stratification layers for sampling design, and structural metrics such as canopy height percentiles, gap fraction, and within-polygon height variability.

\subsection{Limitations}

\paragraph{Temporal and acquisition constraints}
CHMv2 is derived from single-date imagery, where the acquisition process selects the best available image within a target period (2017 -2020). This limits the direct use of the released CHMv2 data for attributing canopy height to a specified year of interest. To support change applications, we provide the image acquisition date associated with each prediction in the dataset metadata.

CHMv2 data users should account for possible local tree detection and height modeling errors due to input image quality issues. Cloud and haze contamination can preclude accurate CHM modeling, and about 10\% of input images were affected by cloud presence. However, cloud masking is imperfect, and residual clouds may further degrade map quality. High off-nadir viewing geometry can also reduce fidelity by inducing map over-generalization and displacement of tree footprints (see example in Figure \ref{fig:plots_multisource}). We estimated that $\approx$20\% of the Maxar images used for this study had off-nadir angles exceeding 25$\degree$. Low sun elevation may further affect CHM prediction. In temperate and boreal regions, low sun angles often correspond to winter-season imagery when deciduous trees are leafless; long shadows cast under these conditions may additionally reduce performance. Analysis of forested samples above 40$\degree$N using ICESat-2 reference data suggests that sun elevation is a strong predictor of map quality. CHMv2 values from input images with sun elevation $<$45$\degree$ show lower agreement ($R^2$=0.42, RMSE=7.4 m) with ICESat-2 than areas with sun elevation $\ge$50$\degree$ ($R^2$=0.6, RMSE=6.5 m). Approximately one-third of the Maxar images used for our map have sun elevations below 45$\degree$, indicating that a substantial portion of CHMv2 coverage in boreal and temperate forests may be affected by suboptimal acquisition conditions.

\paragraph{Gaps and biases in training data}
Although CHMv2 shows improved agreement relative to CHMv1, the geographic distribution of ALS training data is uneven and concentrated in a limited set of geographies. While our validation results demonstrate strong agreement with independent spaceborne LiDAR references (GEDI and ICESat-2), some land cover types and forest structures may exhibit reduced accuracy if they are poorly represented in the training distribution.

\paragraph{Residual artifacts and failure modes}
While CHMv2 improves robustness to input image quality, localized artifacts remain. These primarily include seamlines between neighboring acquisitions, missed haze or thin cloud, and disagreement in areas of extremely tall canopy between different image sources. Despite the significant improvement from CHMv1, CHMv2 still underestimates the upper tail of canopy height distributions, particularly for very tall forests and the crowns of emergent trees. Additionally, because CHMv2 estimates canopy height from a single input image (rather than a multi-date composite), terrain shadow may be a more important cause of measurement error in some regions.

\subsection{Reproducibility statement and environmental impact}

\paragraph{Reproducibility statement}

Our inference code with decoder weights is available in the DINOv3 repository. 
The pseudo code of our gradient loss appears in Algorithm~\ref{alg:gradloss}. 

\begin{algorithm}[t]
\caption{Gradient loss}
\label{alg:gradloss}
\begin{algorithmic}[1]
\State \textbf{Inputs:} prediction $P \in \mathbb{R}^{H\times W}$, target $T \in \mathbb{R}^{H\times W}$, optional mask $M$
\State \textbf{Hyperparams:} small $\varepsilon$, weights $\lambda_{\text{mag}},\lambda_{\text{rng}},\lambda_{\text{dir}}$ (defaults: $0.3,\,0.6,\,0.4$)
\State \textbf{Kernels:} Sobel $G_x, G_y$ \Comment{standard $3\times3$ filters}
\vspace{2pt}
\Function{GradLoss}{$P, T, M$}
  \If{$M$ is not provided} \State $M \gets \mathbf{1}_{H\times W}$ \EndIf
  \State \textbf{(1) Log mean-center}
  \State $\hat{P} \gets \log(\max(P,\varepsilon)) - \mathrm{mean}(\log(\max(P,\varepsilon))\,|\,M)$
  \State $\hat{T} \gets \log(\max(T,\varepsilon)) - \mathrm{mean}(\log(\max(T,\varepsilon))\,|\,M)$

  \State \textbf{(2) Gradients}
  \State $\mathbf{g}_P \gets \big[G_x * \hat{P},\,G_y * \hat{P}\big],\quad \mathbf{g}_T \gets \big[G_x * \hat{T},\,G_y * \hat{T}\big]$
  \State $m_P \gets \|\mathbf{g}_P\|_2,\quad m_T \gets \|\mathbf{g}_T\|_2$

  \State \textbf{(3) Pixel magnitude loss}
  \State $L_{\text{mag}} \gets \mathrm{mean}_M\big[\,|m_P - m_T|\,\big]$

  \State \textbf{(4) 3$\times$3 and 5$\times$5 range matching}
  \State $r_P \gets \mathrm{maxpool}_{3\times3}(m_P) - \mathrm{minpool}_{3\times3}(m_P)$
  \State $r_T \gets \mathrm{maxpool}_{3\times3}(m_T) - \mathrm{minpool}_{3\times3}(m_T)$
  \State $L_{\text{rng}} \gets \mathrm{mean}_M\big[\,|r_P - r_T|\,\big]$

  \State \textbf{(5) direction consistency}
  \State $\hat{\mathbf{u}}_P \gets \mathbf{g}_P/(m_P+\varepsilon),\quad \hat{\mathbf{u}}_T \gets \mathbf{g}_T/(m_T+\varepsilon)$
  \State $c \gets \mathrm{clamp}\!\left(\langle \hat{\mathbf{u}}_P,\hat{\mathbf{u}}_T\rangle,-1,1\right),\quad L_{\text{dir}} \gets \mathrm{mean}_M[\,1-c\,]$

  \State \textbf{(6) Combine}
  \State \Return $L \gets \lambda_{\text{mag}}L_{\text{mag}} + \lambda_{\text{rng}}L_{\text{rng}} + \lambda_{\text{dir}}L_{\text{dir}}$
\EndFunction
\vspace{4pt}
\State \textit{(Optional multi-scale: replace lines 4–24 by a short loop)}
\State \textbf{for} $s \in \{1,\tfrac12,\tfrac14\}$ \textbf{do} downsample $(P,T,M)$ by $s$, compute $L^{(s)}$, accumulate and average.
\end{algorithmic}
\end{algorithm}

\paragraph{Environmental impact}

We estimate the carbon footprint of decoder training using the calculations from \cite{Oquab2023SSL}, with a Thermal Design Power (TDP) of the H100 GPU equal to 700W, a Power Usage Effectiveness (PUE) of 1.1, a carbon intensity factor of 0.385 kg CO$_2$ per KWh, a time of 150 trainings (ablations) $\times$ 3 hours $\times$ 8 GPUs = 3600 GPU hours. The 2772 kWh used to train the model is approximately equivalent to a CO$_2$ footprint of 2772 $\times$ 0.385 = 1.1T of CO$_2$. We estimate the global map inference footprint to 3T of CO$_2$. 

\bibliography{references}

\section{Data Availability} 

The CHMv2 maps are available at \td{\url{link}}.
The 3DEP-NAIP dataset will be made available on Zenodo following manuscript publication. 
The ALS sources of the SatLidar dataset are listed in the DINOv3~\cite{simeoni2025dinov3} paper. This work used the Microsoft Open Buildings dataset\cite{microsoft2018usbuildings}, available under the Open Database License (ODbL) v1.0 license, to clean and curate the NAIP-3DEP dataset.

\section{Code Availability}

The CHMv2 backbone, decoder weights, and example inference code will be available in the DINOv3 GitHub repository.

%

\section{Acknowledgments}

We would like to thank the DINO team, Natacha Supper, Daniel Haziza and Christian Keller, Patrick Nease, and Laura McGorman for their precious help. We would also like to thank past team members for their contribution to this project: Tobias Tiecke, Tracy Johns, Benjamin Nosarzewski, Hung-I Yang, Guillaume Couairon, Sayantan Majumdar, Janaki Vamaraju, Theo Moutakanni and Brian White.

\section{Authors information}

\subsection{Contributions}

\textbf{J.B}: Training data collection, data registration and quality checking, loss design, architecture, sampling changes, analysis of results, paper writing, project coordination. 
\textbf{S.Y.}: GPU Optimization, global Inference of maps, data and OSS preparation.
\textbf{J.T.}: Training data collection, global inference of maps, data processing for comparative analysis.
\textbf{H.V.}: Decoder training with data mix and sampler ablation.
\textbf{M.R.}: Software contributions, loss ablation. 
\textbf{P.L., P.B.}: Technical advice throughout the project. 
\textbf{C.C.}: Data preparation, Decoder training, comparative evaluation, model selection, paper writing (ablation, training sections), project coordination.
\textbf{J.S.}: Analysis of results, paper writing, comparative evaluation.
\textbf{J.E.}: Analysis of results, paper writing.
\textbf{P.P.}: Analysis of results, paper writing, evaluation with spaceborne LiDAR.
\textbf{X.L.}: Analysis of results, evaluation with spaceborne LiDAR.

All authors read and approved the final manuscript.

\subsection{Funding}
The World Resources Institute was supported by a contract from Meta and a grant from the Bezos Earth Fund. Meta and FAIR received no financial support for this research, the preparation of the manuscript, or the publication process. 

\subsection{Competing interests}

The authors declare no competing interests.

\end{document}